\documentclass[journal]{IEEEtran}

\ifCLASSINFOpdf
\else
\fi

\usepackage{amsfonts}
\usepackage{amsmath}
\usepackage{graphicx}
\usepackage{bm}
\usepackage{booktabs}
\usepackage{threeparttable}
\usepackage{multirow}
\usepackage{mathrsfs}
\usepackage{float}
\usepackage{subfigure}
\usepackage{tabularx}
\usepackage{bbding}
\usepackage{tablefootnote}
\usepackage{makecell}
\usepackage{threeparttable}
\usepackage{graphicx}
\usepackage{subfigure}
\usepackage[colorlinks,linkcolor=blue]{hyperref}
\begin{document}

\title{Learning to Incorporate Texture Saliency Adaptive Attention to Image Cartoonization}

\author{Xiang~Gao,~Yuqi~Zhang~,~and~Yingjie~Tian
\thanks{Xiang Gao is with School of Computer Science and Technology, University of Chinese Academy of Sciences, Beijing 100049, China. (e-mail: gaoxiang181@mails.ucas.ac.cn)}
\thanks{Yuqi Zhang is with School of Mathematical Sciences, University of Chinese Academy of Sciences, Beijing 100049, China. (e-mail: zhangyuqi201@mails.ucas.ac.cn)}
\thanks{Yingjie Tian is with Research Center on Fictitious Economy and Data Science, Chinese Academy of Sciences, Beijing 100190, China. (e-mail: tyj@ucas.ac.cn)}} 

\maketitle

\begin{abstract}
Image cartoonization is recently dominated by generative adversarial networks (GANs) from the perspective of unsupervised image-to-image translation, in which an inherent challenge is to precisely capture and sufficiently transfer characteristic cartoon styles (e.g., clear edges, smooth color shading, abstract fine structures, etc.). Existing advanced models try to enhance cartoonization effect by learning to promote edges adversarially, introducing style transfer loss, or learning to align style from multiple representation space. This paper demonstrates that more distinct and vivid cartoonization effect could be easily achieved with only basic adversarial loss. Observing that cartoon style is more evident in cartoon-texture-salient local image regions, we build a region-level adversarial learning branch in parallel with the normal image-level one, which constrains adversarial learning on cartoon-texture-salient local patches for better perceiving and transferring cartoon texture features. To this end, a novel cartoon-texture-saliency-sampler (CTSS) module is proposed to dynamically sample cartoon-texture-salient patches from training data. With extensive experiments, we demonstrate that texture saliency adaptive attention in adversarial learning, as a missing ingredient of related methods in image cartoonization, is of significant importance in facilitating and enhancing image cartoon stylization, especially for high-resolution input pictures. Our code is publically available at \href{https://github.com/XiangGao1102/Learning-to-Incorporate-Texture-Saliency-Adaptive-Attention-to-Image-Cartoonization}{this github link}.
\end{abstract}

\newcommand\blfootnote[1]{
    \begingroup
    \renewcommand\thefootnote{}\footnote{#1}
    \addtocounter{footnote}{-1}
    \endgroup
}
\blfootnote{\textbf{Accepted by ICML 2022}} 


\IEEEpeerreviewmaketitle

\section{Introduction}
\IEEEPARstart{I}{mage} cartoonization aims at rendering natural images into cartoon styles, which is a challenging computer vision and computer graphics problem. Specially designed algorithms that automatically stylize pictures into cartoons can help relieve cartoon creation artists from laborious manual work, and also have practical values in digital entertainment, advertising, childhood education, image editing, etc.

\begin{figure}[t]
\includegraphics[width=3.5in]{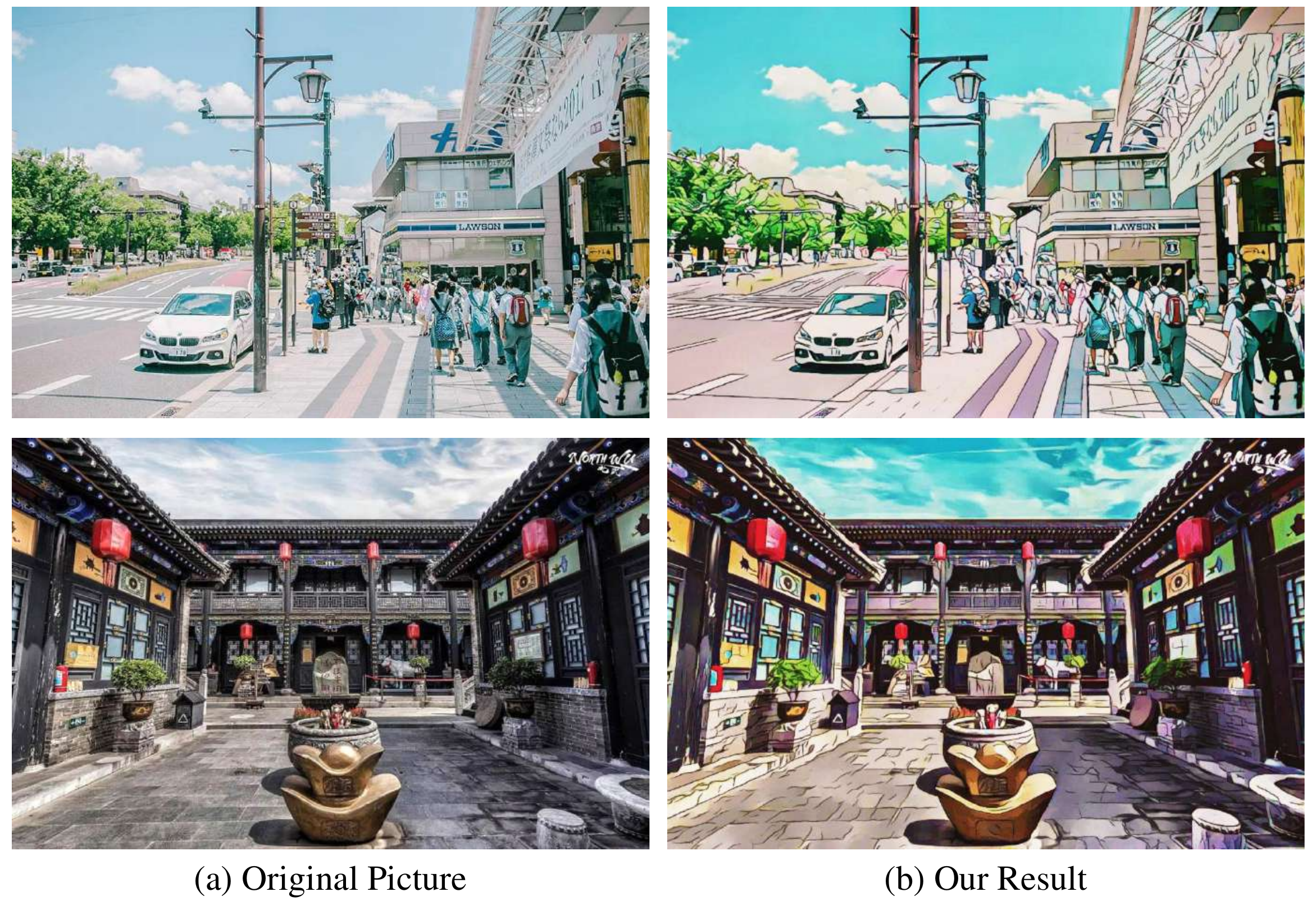}
\caption{Example evaluation results of our method in transforming real-world scenes into cartoon styles (\textbf{better to zoom in}).}
\label{example_results}
\end{figure}
 
Cartoons have unique visual features characterized by clear edges and smooth color shading in non-edge areas. The problem of reproducing cartoon-like effect on real photos is widely explored in early time from the perspective of image abstraction \cite{decarlo2002stylization,winnemoller2006real,kyprianidis2008image,kang2008flow}. These methods model cartoon styles with established image processing techniques, which could be summarized into three aspects: (\romannumeral1) image smoothing; (\romannumeral2) color quantization; (\romannumeral3) edges enhancement. Though successful in mimicking some important cartoon features, they lack data-driven learning ability to capture cartoon styles in more depth.

With the thriving of deep neural networks, recent image cartoonization methods resort to learning-based framework, typically generative adversarial network (GAN) \cite{goodfellow2014generative}, to automatically learn and transfer high-level cartoon styles from real cartoon images. It could be formulated as an unsupervised image-to-image translation problem where the objective is to learn a content-preserving image translation mapping $\mathcal{X}\rightarrow\mathcal{Y}$ from a source domain $\mathcal{X}$ of natural images to a target domain $\mathcal{Y}$ of cartoon images. A general framework is to align style distribution of generated images to that of target-domain real cartoons through adversarial learning, and meanwhile, constrain perceptual content consistency between input photos and generated results to avoid content mismatch. However, it is difficult to produce results with sufficiently salient cartoon features, for which some advanced methods make further progresses to enhance cartoonization effect based on the general framework.

CartoonGAN \cite{chen2018cartoongan} proposes an edge-promoting adversarial loss to highlight typical cartoon feature of edge clearness. This loss function enforces the discriminator to distinguish real cartoon images from not only the synthesized images but also the edge-smoothed cartoon images, such that the generator can be guided to produce clearer edges to fool the discriminator. However, it requires a separate preparation stage to collect an edge-smoothed cartoon image set before training, losing the elegance of end-to-end learning. Furthermore, the edges and contours of the generated results are still not very distinct. AnimeGAN \cite{chen2019animegan} introduces Gram loss \cite{gatys2016image}, a classic texture-descriptor-based style loss widely used in style transfer literature, to GAN framework to enhance learning cartoon texture pattern. Nevertheless, its effect in strengthening cartoon texture transfer is still less noticeable. More recently, a white-box image cartoonization framework \cite{wang2020learning} is proposed. It decomposes images into multiple representations and learns to align styles in the manifold of each representation. This method proposes an adaptive coloring algorithm that generates image color segmentation maps to mimic sparse color blocks of cartoon images, which brings visually appealing celluloid cartoon style. However, the cartoon abstraction and vividness of the generated results are still less prominent, especially for high-resolution input images.

The above-mentioned models resort to different methods to make up for the limitation of the basic adversarial loss in fully transferring cartoon styles. However, we argue that the weak stylization problem is not from the capacity of adversarial loss itself, but due to non-global distribution of salient cartoon texture features. For example, the clear edges are usually distributed in local areas rather than the entire image, and the pixel proportion of clear edges is also very small. Therefore, the feature of edge clearness could be easily overwhelmed by more obvious global features such as color shading smoothness, when trained with basic adversarial loss over the scale of entire image. This enlightens us to attend to cartoon-texture-salient local regions for better perceiving and transferring cartoon texture features.  

To this end, we propose a compact and efficient adversarial learning framework with an image-level discriminator examining global cartoon styles like smooth shading and the overall tone, as well as a patch-level discriminator focusing on learning local cartoon texture pattern, i.e., the unique distribution of high- and low-frequency pixels around clear edges. To enhance transfer of cartoon texture pattern, we adaptively constrain patch-level adversarial learning on cartoon-texture-salient local image regions, for which a novel cartoon-texture-saliency-sampler (CTSS) module is proposed to dynamically extract image patches with most salient cartoon texture pattern from each mini-batch of training images. By incorporating such texture saliency adaptive attention to adversarial learning, the typical cartoon texture features are more sufficiently transferred, yielding more abstract and vivid cartoonization effect. Example results of our model are presented in Fig. \ref{example_results}. Our method bypasses separate edge-smoothing data preparation stage, use of additional style losses, and complicated representation extraction processes, while producing more prominent cartoon effect, especially for large input pictures. The effectiveness of our model is fully demonstrated with extensive experiments evaluated on different cartoon datasets.

\begin{figure*}[t]
\center
\includegraphics[width=7.2in]{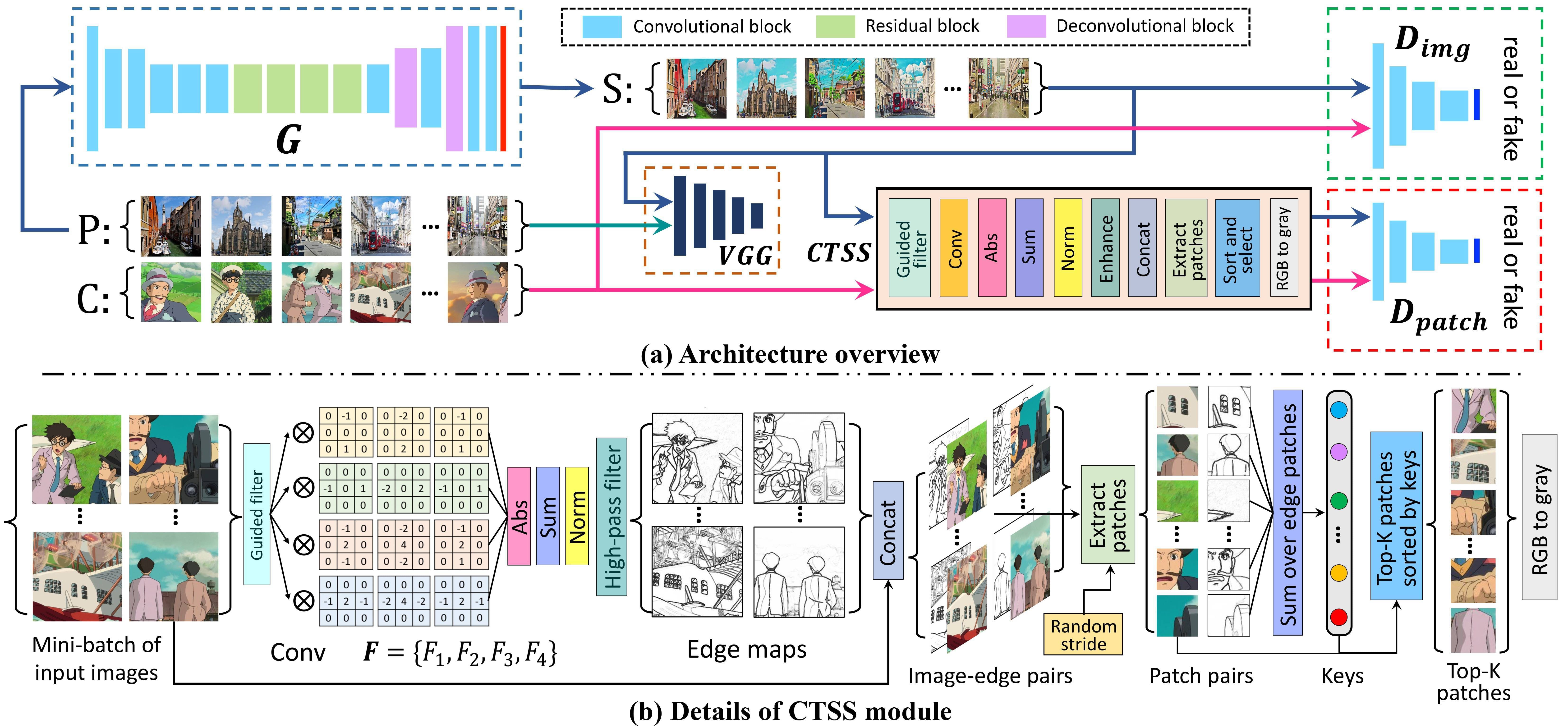}
\caption{The overall architecture of our model, as well as details of our proposed cartoon-texture-saliency-sampler (CTSS) module which adaptively extracts local image patches with most salient cartoon texture pattern from each mini-batch of input images.}
\label{architecture}
\end{figure*}

\section{Related Work}
\subsection{Style Transfer}
Neural style transfer (NST) was first proposed as an online optimization-based algorithm that iteratively transfers image styles by minimizing Gram loss \cite{gatys2016image}. Afterwards, it was transformed into an offline generative model to cater for real-time applications by training a feedforward network \cite{johnson2016perceptual,ulyanov2016texture}. Later on, efforts had been made to extend fast NST from a single style to multiple styles \cite{chen2017stylebank,dumoulin2016learned}, or even arbitrary styles \cite{huang2017arbitrary,li2017universal}. Beside the Gram loss, various style losses were successively proposed, such as MMD loss \cite{li2017demystifying}, CNNMRF loss \cite{li2016combining}, contextual loss \cite{mechrez2018contextual}, and Relaxed EMD loss \cite{kolkin2019style}. These loss functions suit for transferring low-level texture features from a single image. By contrast, cross-domain style transfer methods use adversarial loss to automatically learn high-level styles from a collection of stylistically similar images. By means of GANs, style transfer problem is enriched with more applications, such as font style transfer \cite{tian2017zi2zi,jiang2017dcfont}, painting style transfer \cite{he2018chipgan,gao2020rpd}, makeup style transfer \cite{chang2018pairedcyclegan}, etc.

\subsection{Image-to-Image Translation}
Image-to-image translation refers to transformation of images from a source domain to a target domain, which could be divided into supervised and unsupervised situations according to whether paired training data of two domains are available. For supervised problem, Pix2Pix \cite{isola2017image} combines conditional GAN with image-level sparse $L_{1}$ regularization, which generalizes well to many applications such as image super-resolution \cite{ledig2017photo}, image denoising \cite{alsaiari2019image}, semantic image synthesis \cite{wang2018high}, etc. For the latter situation, CycleGAN \cite{zhu2017unpaired} and UNIT \cite{liu2017unsupervised} are typical models that realize unsupervised image translation based on cycle-consistency constraint and shared-latent-space assumption, respectively. Afterwards, methods like StarGAN \cite{choi2018stargan} and AttGAN \cite{he2019attgan} extend translation from two domains to multiple domains, by combining conditional GAN with auxiliary domain classifier. Besides, the problem is also explored in the field of multimodal translation \cite{huang2018multimodal,lee2018diverse} and few-shot learning \cite{liu2019few}.

\section{Method}
Cartoon images have smooth shading and vivid colors. In addition to these global features that represent overall pixel distribution, the most salient feature of cartoon images is their unique texture pattern that represents local pixel distribution. Specifically, the high-frequency pixels concentrate on edges, while the low-frequency pixels are smoothly distributed beside edges. This distinct separation of high- and low-frequency pixels clearly differs from natural images where the high- and low-frequency elements are much interweaved. However, as Fig. \ref{motivation} shows, such cartoon texture pattern manifests obviously only in partial image regions with clear edges, which means that the latent cartoon texture pattern is more visually perceptible from the view of edge-distinct local regions than from the view of entire image. Therefore, we append a patch-level learning branch that adaptively applies adversarial learning on edge-distinct local regions to enhance capturing cartoon texture pattern.

\subsection{Global and local adversarial learning}
Let $\mathcal{P}$ denote the domain of real-world photos, $\mathcal{C}$ be the domain of cartoon images, $\mathcal{S}$ be the domain of synthesized results. The overall architecture of our adversarial learning framework is illustrated in Fig. \ref{architecture} (a). In training phase, a mini-batch of natural photos $P=\{p_{i}\}_{i=1}^{N} \in \mathcal{P}$, and a mini-batch of cartoon images $C=\{c_{i}\}_{i=1}^{N} \in \mathcal{C}$ are sampled at each iteration, where $N$ is the batch size. A generator $G$ translates $P$ into a mini-batch of synthesized cartoon images $S=G(P)=\{s_{i}\}_{i=1}^{N} \in \mathcal{S}$, which are differentiated from real cartoon images $C$ by an image-level discriminator $D_{img}$. This forms the image-level adversarial learning branch for learning global holistic cartoon styles.

To better seize cartoon texture pattern that is more perceptible at edge-distinct local regions, we append a patch-level adversarial learning branch as a supplement to the image-level one. In this branch, a cartoon-texture-saliency-sampler (CTSS) module is proposed to constrain adversarial learning only on cartoon-texture-distinct local regions. As shown in Fig. \ref{architecture} (a), the CTSS module takes cartoon mini-batch $C$ and the synthesized mini-batch $S$ as input, and outputs top-$K$ edge-distinct local patches $C_{patch}=\{c_{p}^{i}\}_{i=1}^{K}$ and $S_{patch}=\{s_{p}^{i}\}_{i=1}^{K}$ from $C$ and $S$ respectively. A patch-level discriminator $D_{patch}$ is built to distinguish $S_{patch}$ from $C_{patch}$, forming the patch-level adversarial learning that reinforces transfer of cartoon texture pattern.

\subsection{Cartoon-texture-saliency-sampler}
\begin{figure}[t]
\center
\includegraphics[width=3.5in]{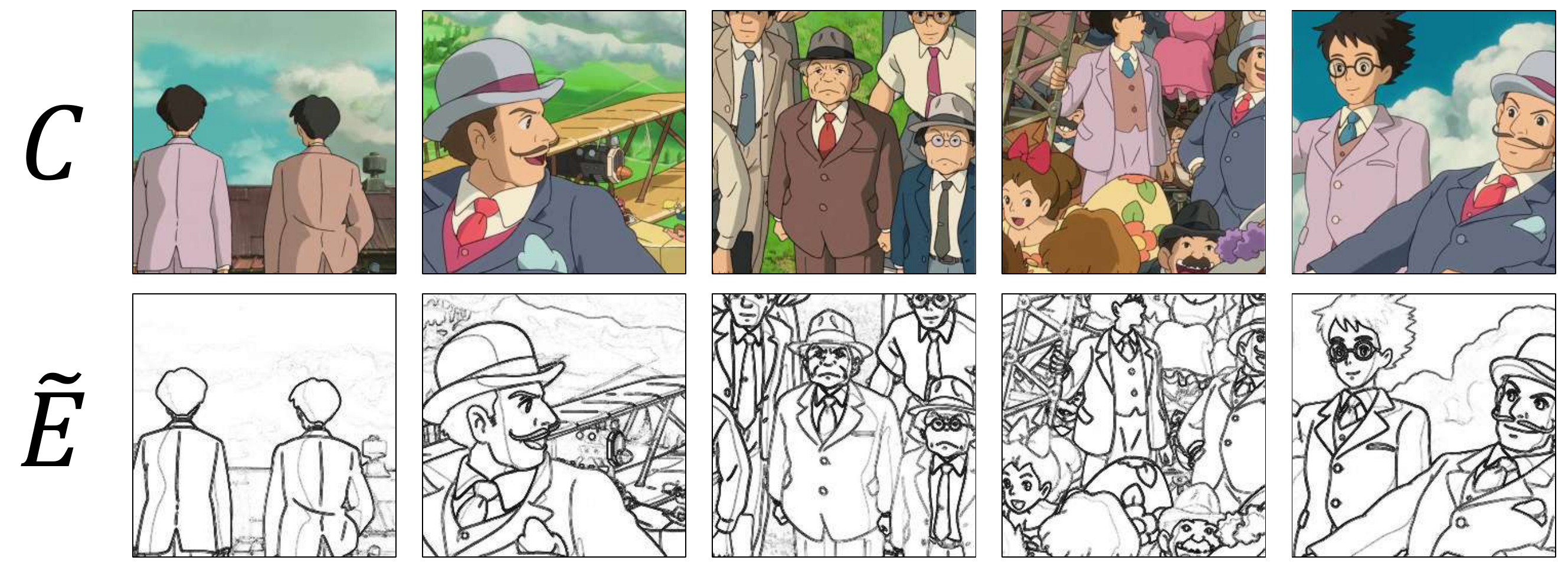}
\caption{Visualization of the refined edge maps $\tilde{E}$ produced during the forward pass of our CTSS module.}
\label{edge_maps}
\end{figure}

Since the unique cartoon texture pattern is more visually prominent at edge-distinct local regions, our CTSS module adaptively samples local image patches with most distinct edges from each input mini-batch of images, the implementation details are illustrated in Fig. \ref{architecture} (b). Taking an input mini-batch of cartoon images $C=\{c_{i}\}_{i=1}^{N}$ as example, CTSS starts with a guided-filter \cite{he2010guided} sub-module $\mathcal{F}_{gf}$ for edge-preserving image smoothing, it uses each input image $c_{i}$ itself as guide map, returns the smoothed image $\tilde{c}_{i}$ with many noise elements removed:
\begin{equation}
\tilde{c}_{i}=\mathcal{F}_{gf}(c_{i}, c_{i}), i=1,...,N.
\label{guided_filter}
\end{equation}
Then, a convolutional layer is applied to extract coarse edge maps $E=\{e_{i}\}_{i=1}^{N}$, where $e_{i}$ is the edge map of $c_{i}$. The convolutional layer has a constant kernel $\textbf{F}$ composed of four filters $\{F_{1},F_{2},F_{3},F_{4}\}$ as shown in Fig. \ref{architecture} (b). The designed kernel $\textbf{F}$ is specially suitable for cartoon edge extraction, and is essentially an improved Sobel operator. The coarse edge map is obtained by summing over the absolute value of the convolution result with each filter of $\textbf{F}$, followed by Min-Max normalization to rescale to $\left[0-1\right]$:
\begin{equation}
e_{i} = Norm_{min\_max}(\sum\nolimits_{k=1}^{4}|\tilde{c}_{i} \otimes F_{k}|), i=1,...,N,
\label{course_edge}
\end{equation}
where $\otimes$ denotes convolution. Eq. \ref{course_edge} can be efficiently implemented with a single-layer convolution with kernel $\textbf{F}$ followed by channel-wise manipulations. Based on coarse edge maps $E$, the refined edge maps $\tilde{E}=\{\tilde{e}_{i}\}_{i=1}^{N}$ are obtained by applying a high-pass filter $h(\cdot)$ that enhances high-frequency pixels and suppresses low-frequency ones:
\begin{equation}
\tilde{e}_{i}=h(e_{i})=1-1/(1+(e_{i}/d)^{n}), i=1,...,N,
\label{high_pass_filter}
\end{equation}
where $d$ and $n$ are hyperparameters that determine threshold and sharpness of the high-pass filter $h(\cdot)$ respectively. Visualization of the final refined edge maps $\tilde{E}$ is shown in Fig. \ref{edge_maps}. The refined edge maps are used to adaptively guide attention to edge-distinct local image regions and extract corresponding image patches:
\begin{figure}[t]
\center
\includegraphics[width=3.5in]{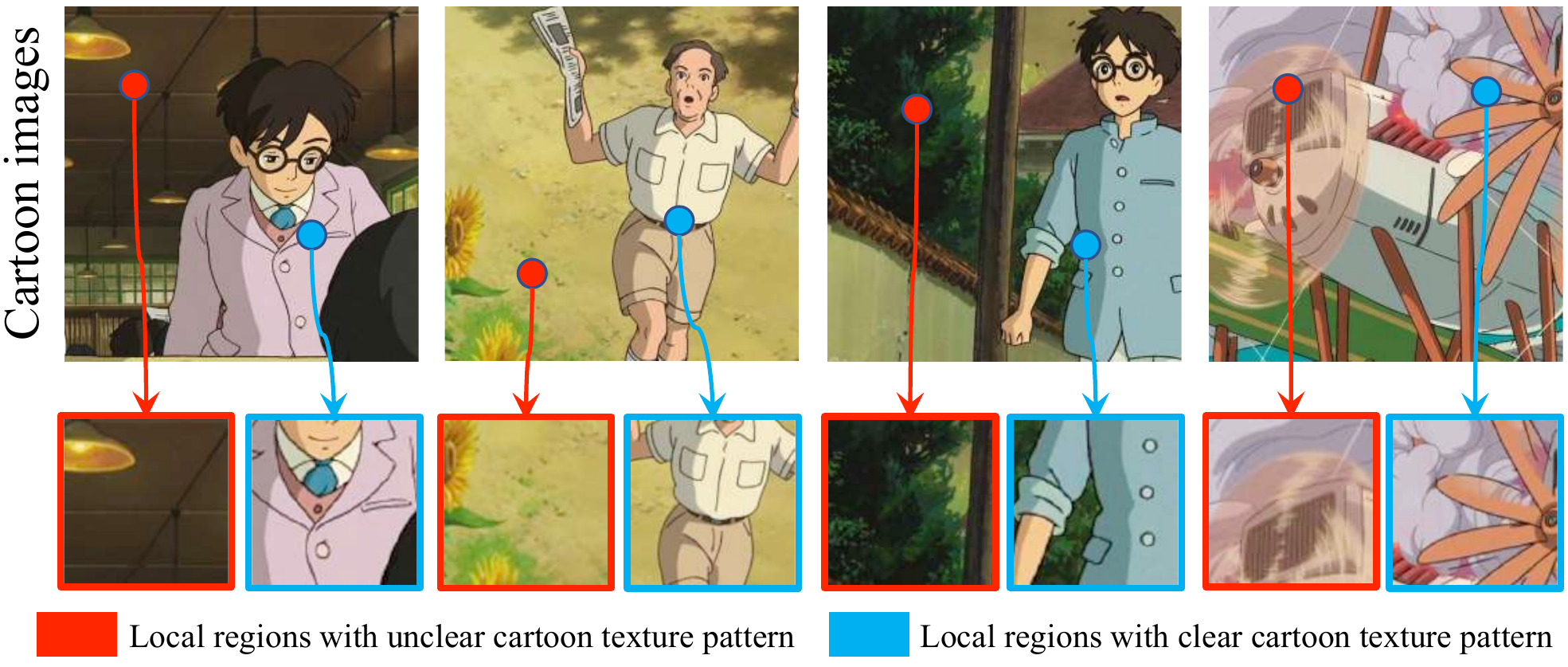}
\caption{The typical cartoon texture pattern manifests clearly only in partial image regions with distinct edges.}
\label{motivation}
\end{figure}
\begin{equation}
\{c_{p}^{i}, e_{p}^{i}\}_{i=1}^{M}=ExtractPatches(C\oplus\tilde{E}, l, s),
\label{patch_extract}
\end{equation}
where $\oplus$ denotes channel-wise concat operation, $l$ and $s$ are respectively patch size and stride of the sliding-window-alike patch extraction process. $c_{p}^{i}$ and $e_{p}^{i}$ are the $i$th extracted patch of cartoon images $C$ and edge maps $\tilde{E}$ respectively, they are paired and correspond to the same image location. $M$ is the total number of the extracted patches from a mini-batch of training images, i.e., $M=N(\lfloor\frac{H-l}{s}\rfloor+1)(\lfloor\frac{W-l}{s}\rfloor+1)$, where $H$ and $W$ are height and width of training images, and $N$ is the batch size. The extracted $M$ image patches are sorted by the edge intensity of their paired edge patches, where the edge intensity is quantified as the pixel summation over an edge patch. After sorting, top-$K$ image patches $C_{patch}$ with most distinct edges are sampled:
\begin{equation}
t_{i}=\sum\nolimits_{m,n}(e_{p}^{i})_{m,n}, i=1,2,...,M,
\end{equation}
\begin{equation}
C_{patch}=\{c_{p}^{a_{j}}, j=1,...,K|t_{a_{1}}\geq t_{a_{2}}\geq\cdot\cdot\cdot\geq t_{a_{M}}\},
\end{equation}

where $m$ and $n$ index pixel coordinate of each local patches, $\{a_{1},a_{2},...,a_{M}\}$ is a permutation of $\{1,2,...,M\}$. The sampled top-$K$ image patches $C_{patch}$ contain most evident cartoon texture pattern, they serve as training data of the patch-level adversarial learning branch to promote learning cartoon texture feature. Considering that local patches can not reflect overall color distribution of images, we finally convert the sampled $C_{patch}$ to grayscale for purpose of learning color-invariant local cartoon texture pattern. Since we use gradient based filtering method to detect edges, some local patches with no clear edges but a lot of noise can still have large accumulated gradients and thus be sampled out for patch-level training. Consequently, we apply guided filtering (Eq. \ref{guided_filter}) for edge-preserving image denoising before edge extraction (Eq. \ref{course_edge}), this guarantees that image patches are sampled out due to clear edges instead of large noises.

It is worth mentioning that our method is more suitable than using pre-trained deep model to detect edges. Firstly, well-trained deep edge detection models tend to make high-confidence prediction to any edge pixels, the generated edge maps are less able to reflect edge intensity difference, and thus can not locate true edge-salient local regions. Secondly, our filtering based method uses only single-layer convolution to produce edge maps, which is much faster than forward propagation through pre-trained deep models.

\subsection{Objective functions}
The training of our model comprises five loss functions, they are content loss, global adversarial loss, local adversarial loss, color reconstruction loss, and total variation loss.

\noindent\textbf{Content loss} is used to guarantee content consistency between input photos and cartoonized results, which is realized by matching feature maps at the $l$th layer of the pre-trained VGG19 \cite{simonyan2014very} network:
\begin{equation}
L_{con}=\mathbb{E}_{p_{i}\sim\mathcal{P}}[||VGG_{l}(p_{i})-VGG_{l}(G(p_{i}))||_{1}],
\label{content_loss}
\end{equation}
where the $l$th layer is ``conv4-4'' in VGG19. 

\noindent\textbf{Global adversarial loss} aims to capture global cartoon style through image-level adversarial learning branch. We employ LSGAN \cite{mao2017least} loss for better stability:
\begin{equation}
L_{adv\_global}=L_{adv\_global}^{D}+L_{adv\_global}^{G},
\label{global_adv_loss}
\end{equation}
\begin{equation}
\begin{aligned}
L_{adv\_global}^{D}=&\mathbb{E}_{c_{i}\sim\mathcal{C}}[(D_{img}(c_{i})-1)^{2}]+\\
            &\mathbb{E}_{p_{i}\sim\mathcal{P}}[(D_{img}(G(p_{i})))^{2}],
\end{aligned}
\end{equation}
\begin{equation}
L_{adv\_global}^{G}=\mathbb{E}_{p_{i}\sim\mathcal{P}}[(D_{img}(G(p_{i}))-1)^{2}].
\end{equation}

\noindent\textbf{Local adversarial loss} aims at learning local cartoon texture pattern through patch-level adversarial learning branch:
\begin{equation}
L_{adv\_local}=L_{adv\_local}^{D}+L_{adv\_local}^{G},
\label{local_adv_loss}
\end{equation} 
\begin{equation}
\begin{aligned}
L_{adv\_local}^{D}=&\mathbb{E}_{C_{patch}}[\frac{1}{K}\sum\nolimits_{i=1}^{K}(D_{patch}(c_{p}^{i})-1)^{2}]\\
           +&\mathbb{E}_{S_{patch}}[\frac{1}{K}\sum\nolimits_{i=1}^{K}(D_{patch}(s_{p}^{i}))^{2}], 
\end{aligned}
\end{equation}
\begin{equation}
L_{adv\_local}^{G}=\mathbb{E}_{S_{patch}}[\frac{1}{K}\sum\nolimits_{i=1}^{K}(D_{patch}(s_{p}^{i})-1)^{2}],
\end{equation}

where $C_{patch}=\{c_{p}^{i}\}_{i=1}^{K}$ are extracted top-$K$ edge-distinct patches from $C=\{c_{i}\}_{i=1}^{N}$, $S_{patch}=\{s_{p}^{i}\}_{i=1}^{K}$ are top-$K$ edge-distinct patches from $S=\{s_{i}\}_{i=1}^{N}=\{G(p_{i})\}_{i=1}^{N}$.

\noindent\textbf{Color reconstruction loss} is used to retain color information after cartoonization. Following \cite{chen2019animegan}, we convert image from RGB to YUV format, and apply $L_{1}$ loss to Y channel and Huber Loss to U and V channels:
\begin{equation}
\begin{aligned}
L_{col}=\mathbb{E}_{p_{i}\sim\mathcal{P}}[||Y(G(p_{i}))-Y(p_{i})||_{1}+\\
||U(G(p_{i}))-U(p_{i})||_{H}+||V(G(p_{i}))-V(p_{i})||_{H}],
\end{aligned}
\end{equation}
where $Y(\cdot)$, $U(\cdot)$, $V(\cdot)$ represent the three channels of an image in YUV format, and $H$ denotes Huber Loss.

\noindent\textbf{Total variation loss} is used to reduce noises and artifacts of the generated results:
\begin{equation}
\begin{aligned}
L_{tv}=&\mathbb{E}_{s_{i}\sim\mathcal{S}}[\frac{1}{H(W-1)}\sum_{r=1}^{H}\sum_{c=1}^{W-1}(s_{i_{r,c+1}}-s_{i_{r,c}})^{2}\\
+&\frac{1}{(H-1)W}\sum_{r=1}^{H-1}\sum_{c=1}^{W}(s_{i_{r+1,c}}-s_{i_{r,c}})^{2}],
\end{aligned}
\end{equation}
where $H$ and $W$ are height and width of generated images. 

\noindent The total loss function can be decomposed into a generator part and a discriminator part:
\begin{equation}
\begin{aligned}
L_{gen}=&\lambda_{global}L_{adv\_global}^{G}+\lambda_{local}L_{adv\_local}^{G}+\\
        &\lambda_{con}L_{con}+\lambda_{col}L_{col}+\lambda_{tv}L_{tv},
\end{aligned}
\label{generator_loss}
\end{equation}
\begin{equation}
L_{dis}=\lambda_{global}L_{adv\_global}^{D}+\lambda_{local}L_{adv\_local}^{D},
\label{discriminator_loss}
\end{equation}

where $L_{gen}$ is minimized to optimize the generator $G$, $L_{dis}$ is minimized to jointly optimize the two discriminators $D_{img}$ and $D_{patch}$. $L_{gen}$ and $L_{dis}$ are minimized alternately to form the adversarial training framework.

\section{Experiments}
\subsection{Datasets}

Our model can be easily trained with unpaired data. The source-domain real-scene photos comprise 6656 images for training, 790 images for quantitative testing, and 300 high-resolution images for qualitative evaluation. The training and quantitative testing sets are borrowed from the training and testing sets of CycleGAN \cite{zhu2017unpaired} respectively, where the spatial size of all images are fixed at 256$\times$256 for fair comparison with baseline models \cite{chen2018cartoongan, chen2019animegan, wang2020learning}. We additionally collect 300 high-resolution pictures with width ranging from 960 to 3000 pixels to qualitatively evaluate models' cartoonization effect on practical large input pictures. For target-domain training data, we prepare three cartoon datasets of different styles. They are respectively consisted of cartoon frames cropped from ``The Wind Rises'', ``Dragon Ball'', and ``Crayon Shin-chan''. Each cartoon dataset has 2000 cartoon images, they are rescaled to 256$\times$256 to be in accordance with the size of source domain training images.
\subsection{Parameter settings}
The training batch size is $N$=8, the number of local patches extracted from each mini-batch is $K$=32. We set $d$=0.2, $n$=2 in Eq. \ref{high_pass_filter}, which empirically has good effect in refining edge maps. For patch extraction, we set patch size $l$=96, and randomly sample moving stride $s$ in Eq. \ref{patch_extract} from a uniform distribution $U(48,72)$ at each training iteration, as a kind of patch-level data augmentation strategy.

\begin{figure*}[t]
\center
\includegraphics[width=7.1in]{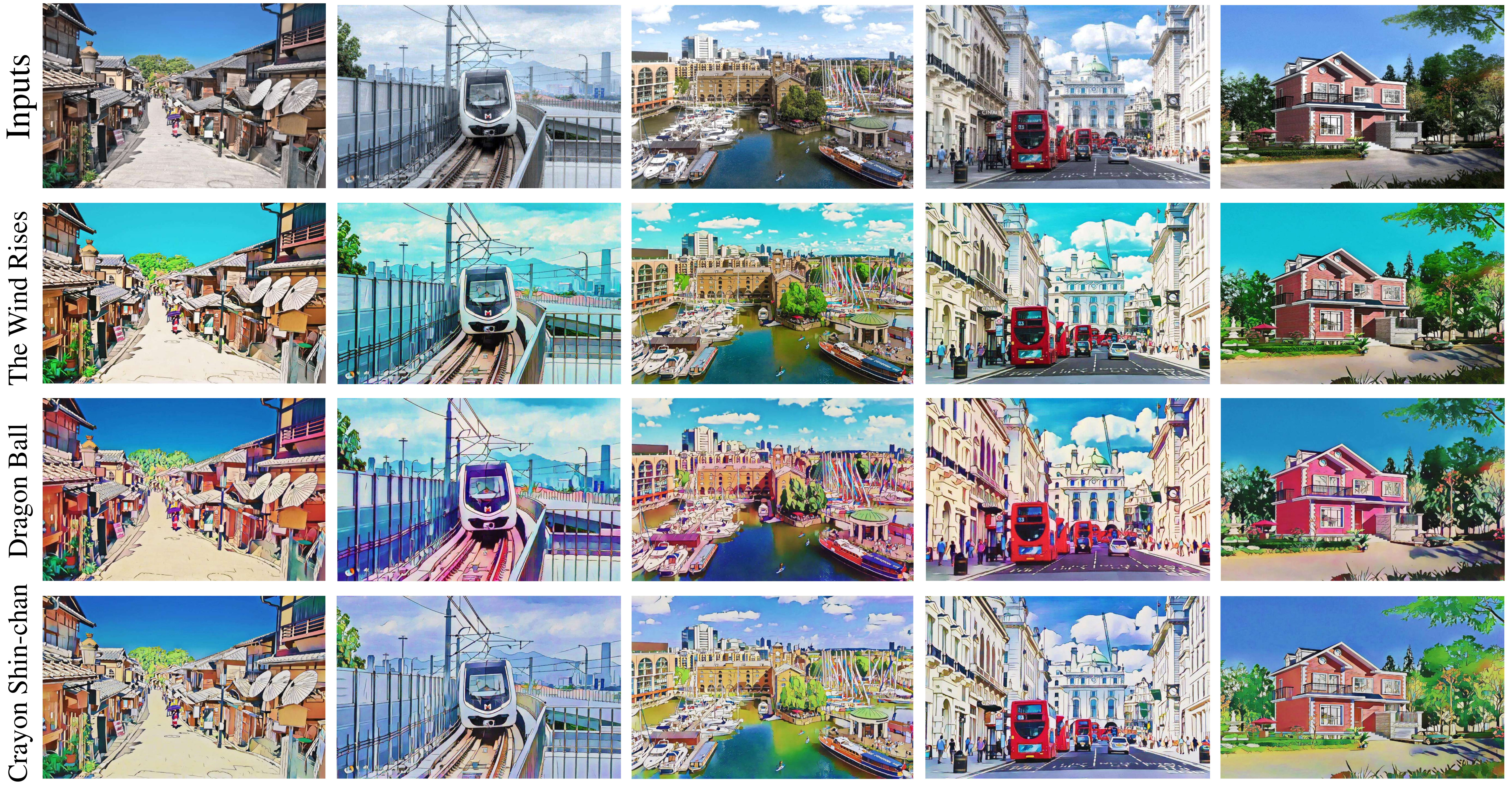}
\caption{Example image cartoonization results tested over high-resolution real-world-scene input images. Results are evaluated on our model trained over different cartoon datasets, including ``The Wind Rises'' (the second row), ``Dragon Ball'' (the third row), and ``Crayon Shin-chan'' (the bottom row). \textbf{Please zoom in for better resolution}.}
\label{qualitative_results}
\end{figure*}

\begin{figure*}[t]
\center
\includegraphics[width=7.1in]{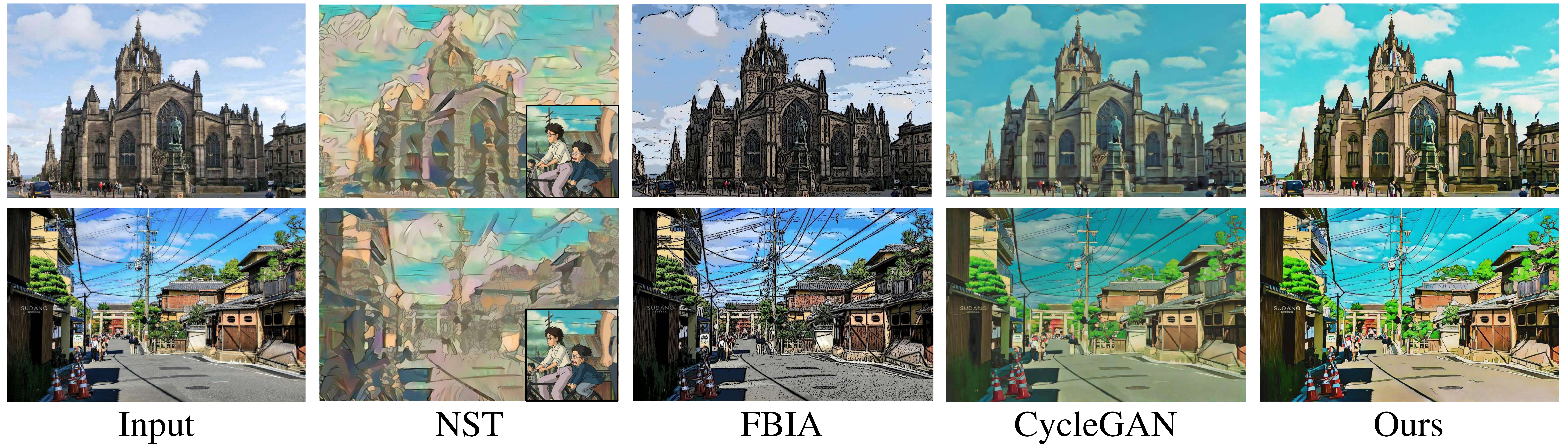}
\caption{Comparison of our approach with general image stylization or abstraction methods tested over high-resolution real-world-scene input images (\textbf{better to zoom in}). Results are evaluated on ``The Wind Rises'' dataset.}
\label{method_compare_1}
\end{figure*}
\begin{figure*}[h!]
\center
\includegraphics[width=7.1in]{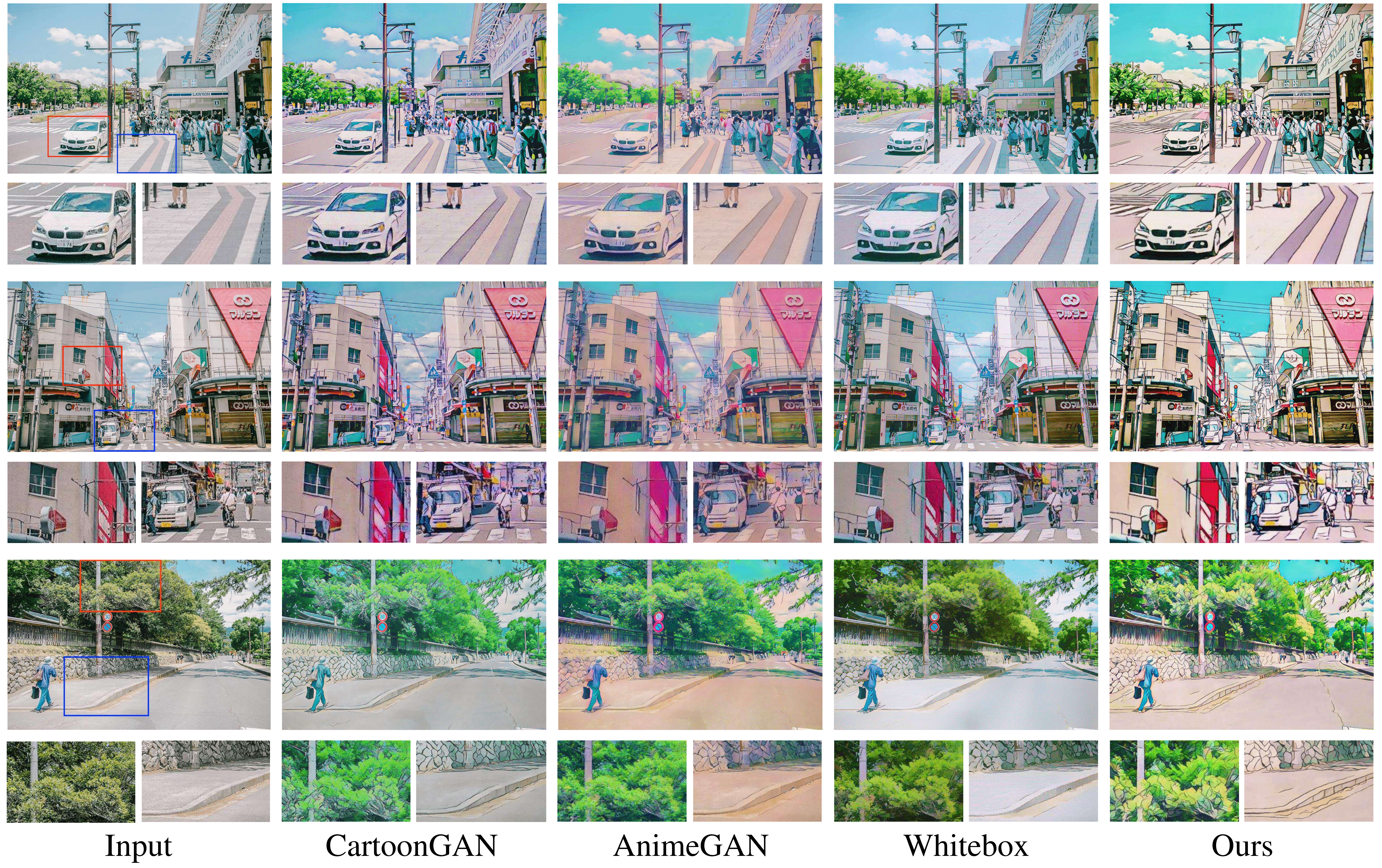}
\caption{Comparison of our approach with related advanced image cartoonization methods tested over high-resolution real-world-scene input images (\textbf{better to zoom in for details}). Results are evaluated on ``The Wind Rises'' dataset. For all related methods, the image natural saturation of the generated results is increased by 40\% as well for fair comparison with our method.}
\label{method_compare_2}
\end{figure*}

\subsection{Training and inference details}
At training phase, we set the weights of component loss functions to be $\lambda_{global}$=$\lambda_{local}$=300$, \lambda_{con}$=1.5, $\lambda_{col}$=15, $\lambda_{tv}$=1. We train our model for 80 epochs. In the first 10 epochs, we pre-train the generator by minimizing $L_{con}$ (Eq. \ref{content_loss}) with an initial learning rate of $2\times 10^{-4}$. In the remaining 70 epochs, we alternately minimize $L_{gen}$ (Eq. \ref{generator_loss}) and $L_{dis}$ (Eq. \ref{discriminator_loss}) to optimize the generator part and the discriminator part respectively, where the initial learning rate for both two parts are $2\times 10^{-5}$. We use Adam optimizer with $\beta_{1}$=0.5, $\beta_{2}$=0.999. At inference time, we dynamically increase image natural saturation of the generated results by 40\% to produce more vivid colors.
\subsection{Qualitative and quantitative results}

Some qualitative results of our model trained over different cartoon datasets and evaluated over high-resolution input pictures are shown in Fig. \ref{qualitative_results}, our method reproduces vivid cartoon effects on high-resolution real-world-scene photos. For qualitative method comparison, we divide related methods into general image stylization or abstraction methods including neural style transfer (NST) \cite{gatys2016image}, flow-based image abstraction (FBIA) \cite{kang2008flow}, and CycleGAN \cite{zhu2017unpaired}, as well as advanced image cartoonization methods including CartoonGAN \cite{chen2018cartoongan}, AnimeGAN \cite{chen2019animegan}, and WhiteBox \cite{wang2020learning}. Results of these models trained over ``The Wind Rises'' dataset are shown in Fig. \ref{method_compare_1} and Fig. \ref{method_compare_2}. NST globally and randomly transfers low-level texture features, the produced results suffer from unappealing artifacts and structure distortions. FBIA abstracts images with learning-free image processing techniques, the generated results fail to present the distribution of target cartoon style. Results of CycleGAN do not manifest local cartoon texture features such as clear edges.

\begin{figure*}[t]
\center
\includegraphics[width=7.1in]{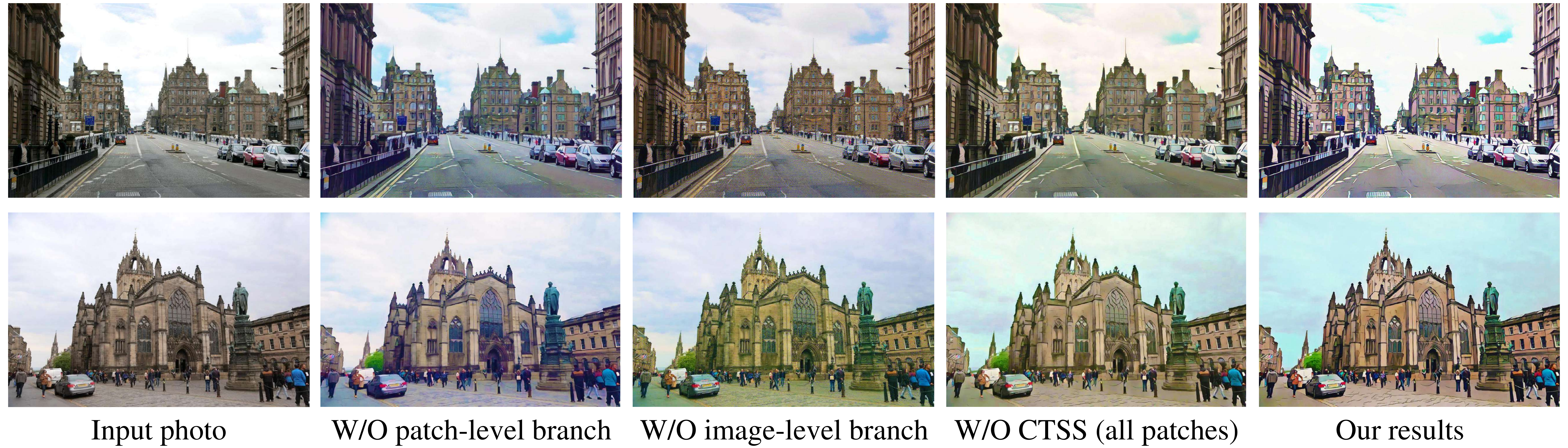}
\caption{Qualitative ablation study of the components of our framework, including the image-level learning branch, patch-level learning branch, and our CTSS module (\textbf{better to zoom in}). Results are evaluated on our model trained over ``The Wind Rises'' dataset.}
\label{ablation_study}
\end{figure*}

\begin{figure*}[t]
\center
\includegraphics[width=7.2in]{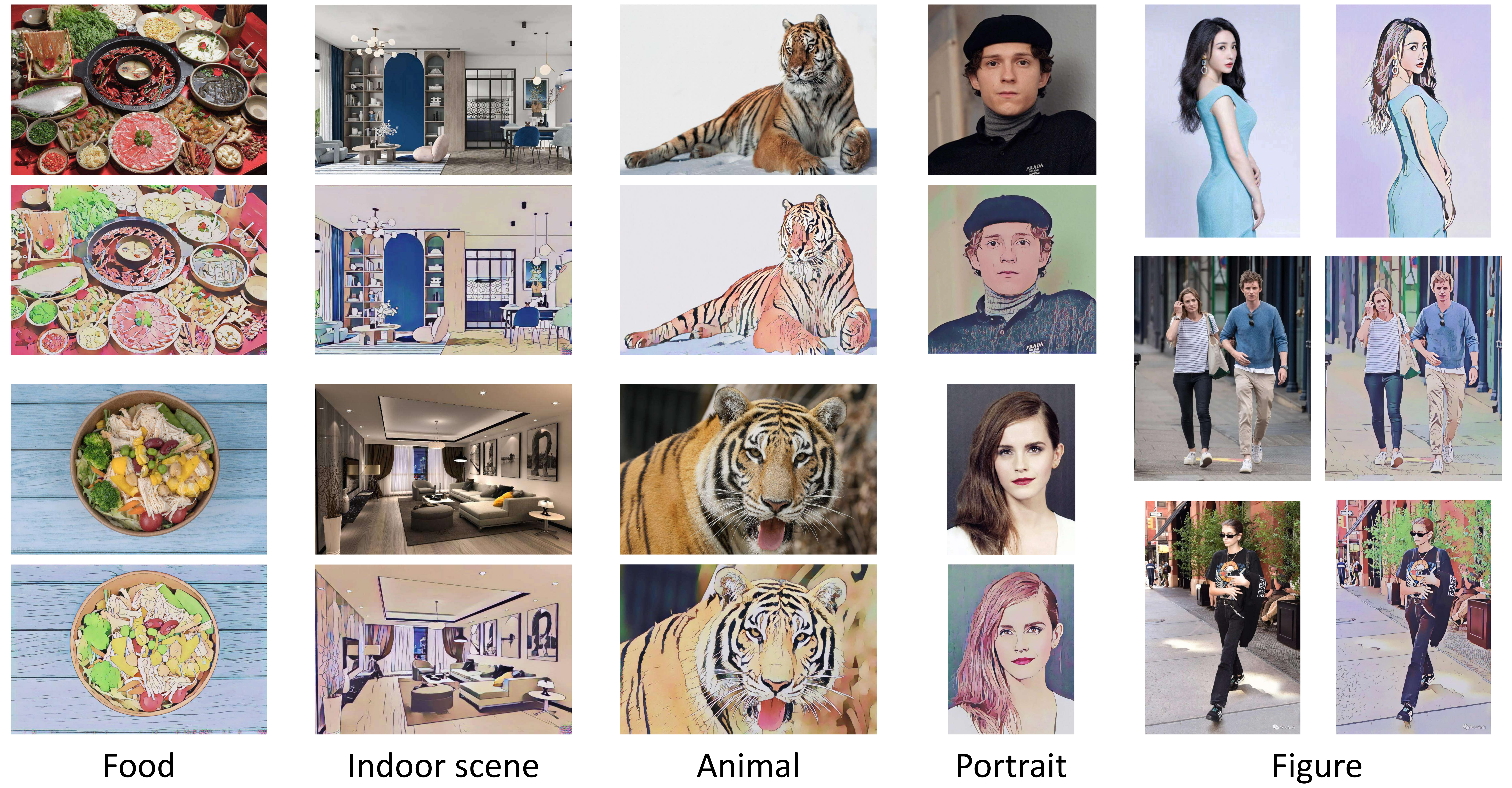}
\caption{Qualitative cartoonization results evaluated on ``Crayon Shin-chan'' dataset in more scenarios including foods, indoor scenes, animals, portraits, and figures. \textbf{Better to zoom in for higher resolution}.}
\label{more_scenarios}
\end{figure*}

For all the advanced image cartoonization methods to be compared with, we keep the same training epochs as our method. As Fig. \ref{method_compare_2} demonstrates, the stylization degree of CartoonGAN and AnimeGAN is relatively weak, the typical local cartoon texture pattern is not evident in generated results. Note that both these two methods prepare an edge-smoothed cartoon dataset to explicitly learn to promote edges adversarially, the generated edges are still not as clear as ours. Results of WhiteBox do not exhibit much cartoon abstraction and vividness for high-resolution input images. All these advanced methods are not able to fully capture the inherent cartoon texture features. By contrast, our results present prominent cartoon texture pattern, and sufficient abstraction and vividness. Apart from landscape pictures, results of our method evaluated on ``Crayon Shin-chan'' dataset in more scenarios are displayed in Fig. \ref{more_scenarios}.

We quantitatively evaluate model performance by measuring the FID between the collection of generated images and the collection of real cartoon images. Results of FID evaluated on the source-domain test set are reported in Tab. \ref{tab:FID_comparison}. Our method achieves much lower FID than related GAN-based methods, indicating that the style distribution of our results is much closer to that of target-domain real cartoon images. Besides, we investigate the influence of $K$ (i.e., the total number of local patches sampled from each mini-batch of input images) to model performance. Results reported in Tab. \ref{tab:investigation_of_K} show that both too small and too large value of $K$ may degrade model performance. This is because sampling less patches leads to training insufficiency of patch-level adversarial learning, while oversampling patches results in many low-quality patches with less salient cartoon textures, which weakens our model's ability to capture and highlight cartoon texture features.

\renewcommand{\arraystretch}{1.8}
\begin{table}[t]
  \begin{threeparttable}
  \centering
  \fontsize{9}{8}\selectfont
  \caption{Comparison with related methods in FID $\downarrow$.}
  \label{tab:FID_comparison}
    \begin{tabular}{|p{29mm}<{\centering}|p{14mm}<{\centering}|p{14mm}<{\centering}|p{14mm}<{\centering}|}
    \hline
    \multirow{2}{*}{Models}&
    \multicolumn{3}{c|}{FID}\cr\cline{2-4}
    &TWR&DB&CSC\cr
    \hline
    \hline
    CycleGAN&146.45&141.33&142.86\cr\hline
    CartoonGAN&143.96&145.58&147.36\cr\hline
    AnimeGAN&136.12&134.94&138.83\cr\hline
    WhiteBox&132.67&137.45&140.79\cr\hline
    Ours&{\bf115.25}&{\bf112.97}&{\bf124.66}\cr\hline
    \end{tabular}
    \begin{tablenotes}
	\item TWR, DB, CSC respectively denote ``The Wind Rises'', ``Dragon Ball'', and ``Crayon Shin-chan'' dataset.
	\end{tablenotes}
	\end{threeparttable}
\end{table}

\renewcommand{\arraystretch}{1.8}
\begin{table}[t]
  \begin{threeparttable}
  \centering
  \fontsize{9}{8}\selectfont
  \caption{Influence of $K$ to FID $\downarrow$.}
  \label{tab:investigation_of_K}
    \begin{tabular}{|p{14mm}<{\centering}|p{13mm}<{\centering}|p{13mm}<{\centering}|p{13mm}<{\centering}|p{13mm}<{\centering}|}
    \hline
    \multirow{2}{*}{Dataset}&
    \multicolumn{4}{c|}{FID}\cr\cline{2-5}
    &$K$=16&$K$=32&$K$=64&$K$=128\cr
    \hline
    \hline
    TWR&115.86&{\bf115.25}&116.47&126.78\cr\hline
    DB&113.53&{\bf112.97}&113.37&125.64\cr\hline
    CSC&125.45&{\bf124.66}&124.94&130.33\cr\hline
    \end{tabular}
    \begin{tablenotes}
	\item TWR, DB, and CSC denote ``The Wind Rises'', ``Dragon Ball'', and ``Crayon Shin-chan'' cartoon dataset respectively.
	\end{tablenotes}
	\end{threeparttable}
\end{table}

\renewcommand{\arraystretch}{1.8}
\begin{table}[t]
  \begin{threeparttable}
  \centering
  \fontsize{9}{8}\selectfont
  \caption{Ablation study of our model with FID $\downarrow$.}
  \label{tab:ablation_study}
    \begin{tabular}{|p{35mm}<{\centering}|p{12mm}<{\centering}|p{12mm}<{\centering}|p{12mm}<{\centering}|}
    \hline
    \multirow{2}{*}{Models}&
    \multicolumn{3}{c|}{FID}\cr\cline{2-4}
    &TWR&DB&CSC\cr
    \hline
    \hline
    W/O patch-level branch&145.25&146.63&149.84\cr\hline
    W/O image-level branch&182.78&185.66&190.04\cr\hline
    W/O CTSS (all patches)&127.09&125.80&130.90\cr\hline
    Ours (full model)&{\bf115.25}&{\bf112.97}&{\bf124.66}\cr\hline
    \end{tabular}
	\end{threeparttable}
\end{table}

\renewcommand{\arraystretch}{1.8}
\begin{table}[t]
  \begin{threeparttable}
  \centering
  \fontsize{9}{8}\selectfont
  \caption{Ablation study of loss functions for stylization.}
  \label{tab:loss_study}
    \begin{tabular}{|p{46mm}<{\centering}|p{8mm}<{\centering}|p{8mm}<{\centering}|p{8mm}<{\centering}|}
    \hline
    \multirow{2}{*}{Loss functions for stylization}&
    \multicolumn{3}{c|}{FID}\cr\cline{2-4}
    &TWR&DB&CSC\cr
    \hline
    \hline
    $L_{adv\_global}$&145.25&146.63&149.84\cr\hline
    $L_{ep\_adv\_global}$&143.40&145.64&147.75\cr\hline
    $L_{adv\_global}+L_{gram}$&139.05&137.71&141.66\cr\hline
    $L_{ep\_adv\_global}+L_{gram}$&137.43&135.50&140.18\cr\hline
    $L_{ep\_adv\_global}+L_{mean\_std}$&135.99&135.24&137.54\cr\hline
    $L_{ep\_adv\_global}+L_{gram} + L_{mean\_std}$&135.45&134.60&136.08\cr\hline
    $L_{adv\_global}+L_{adv\_local}$ (ours)&{\bf115.25}&{\bf112.97}&{\bf124.66}\cr\hline
    \end{tabular}
    \begin{tablenotes}
	\item $L_{ep\_adv\_global}$ denotes the edge-promoting adversarial loss proposed in CartoonGAN \cite{chen2018cartoongan}. 
	\item $L_{gram}$ and $L_{mean\_std}$ are Gram loss \cite{gatys2016image} and mean-variance loss \cite{huang2017arbitrary} respectively.
	\end{tablenotes}
	\end{threeparttable}
\end{table}

\subsection{Ablation study}
Ablation studies of the components of our model are qualitatively and quantitatively shown in Fig. \ref{ablation_study} and Tab. \ref{tab:ablation_study} respectively. Removing patch-level adversarial learning branch leads to results with weak cartoon styles and large FID, reflecting that image-level adversarial loss alone is not sufficient to transfer salient cartoon texture features. When removing the image-level learning branch, we observe that the model with only patch-level branch suffers from training instability. The adversarial learning fails to converge and the corresponding results do not exhibit any cartoon styles. This indicates the fundamental role of image-level branch in maintaining the balance of adversarial training. Combining both two branches constitutes our full model that produces visually appealing cartoon effects with prominent cartoon texture pattern and substantially lower FID. Lastly, to verify that the performance gain stems from not only patch-level adversarial learning but also adaptive texture saliency local attention brought by our CTSS module, we remove CTSS module and extract all local patches from each mini-batch of images for patch-level adversarial learning, the corresponding results are noticeably inferior to our model with CTSS module both qualitatively and quantitatively. This indicates that adaptively constrain patch-level adversarial learning on cartoon-texture-salient local regions is indeed contributive to enhancing cartoon style rendering.

Besides, we investigate the effectiveness of style related loss functions in image cartoonization, results are reported in Tab. \ref{tab:loss_study}. Compared with the basic image-level adversarial loss $L_{adv\_global}$ (Eq. \ref{global_adv_loss}), the edge-promoting adversarial loss $L_{ep\_adv\_global}$ used in CartoonGAN \cite{chen2018cartoongan} brings very limited performance gains at the cost of a prior edge-smoothing data preparation stage. Following the idea of AnimeGAN \cite{chen2019animegan} to introduce style transfer losses into GAN framework, we combine  the image-level adversarial loss $L_{adv\_global}$ or $L_{ep\_adv\_global}$ with the second-order style transfer loss $L_{gram}$ \cite{gatys2016image} or the first-order one  $L_{mean\_std}$ \cite{huang2017arbitrary}. Results show that introduction of style transfer losses is useful to narrow style distribution gap between generated results and real cartoons. However, these extra style losses transfer global and low-level image styles \cite{li2017laplacian}, not able to fully capture local cartoon texture pattern. By contrast, our method improves stylization performance dramatically simply by applying the basic adversarial loss at local views, which bypasses the need of prior data preprocessing stage and additional style transfer losses.

\section{Conclusion}
This paper proposes an end-to-end deep generative model for image cartoonization. We supplement the normal image-level adversarial learning with a patch-level learning branch, and adaptively constrain patch-level adversarial learning only on cartoon-texture-salient local regions to enhance capturing typical cartoon texture pattern. To this end, a novel cartoon-texture-saliency-sampler (CTSS) module is proposed to dynamically sample local image patches containing most prominent cartoon texture features. By incorporating such texture saliency adaptive attention to adversarial learning, our method is able to transfer noticeably more abstract and vivid cartoon styles than related methods.

\section{Acknowledgments}
This work has been partially supported by grants from: National Natural Science Foundation of China (No.12071458, 71731009)

\bibliographystyle{IEEEtran}
\bibliography{mybib}

\newpage

\begin{figure*}[h!]
\center
\includegraphics[width=7in]{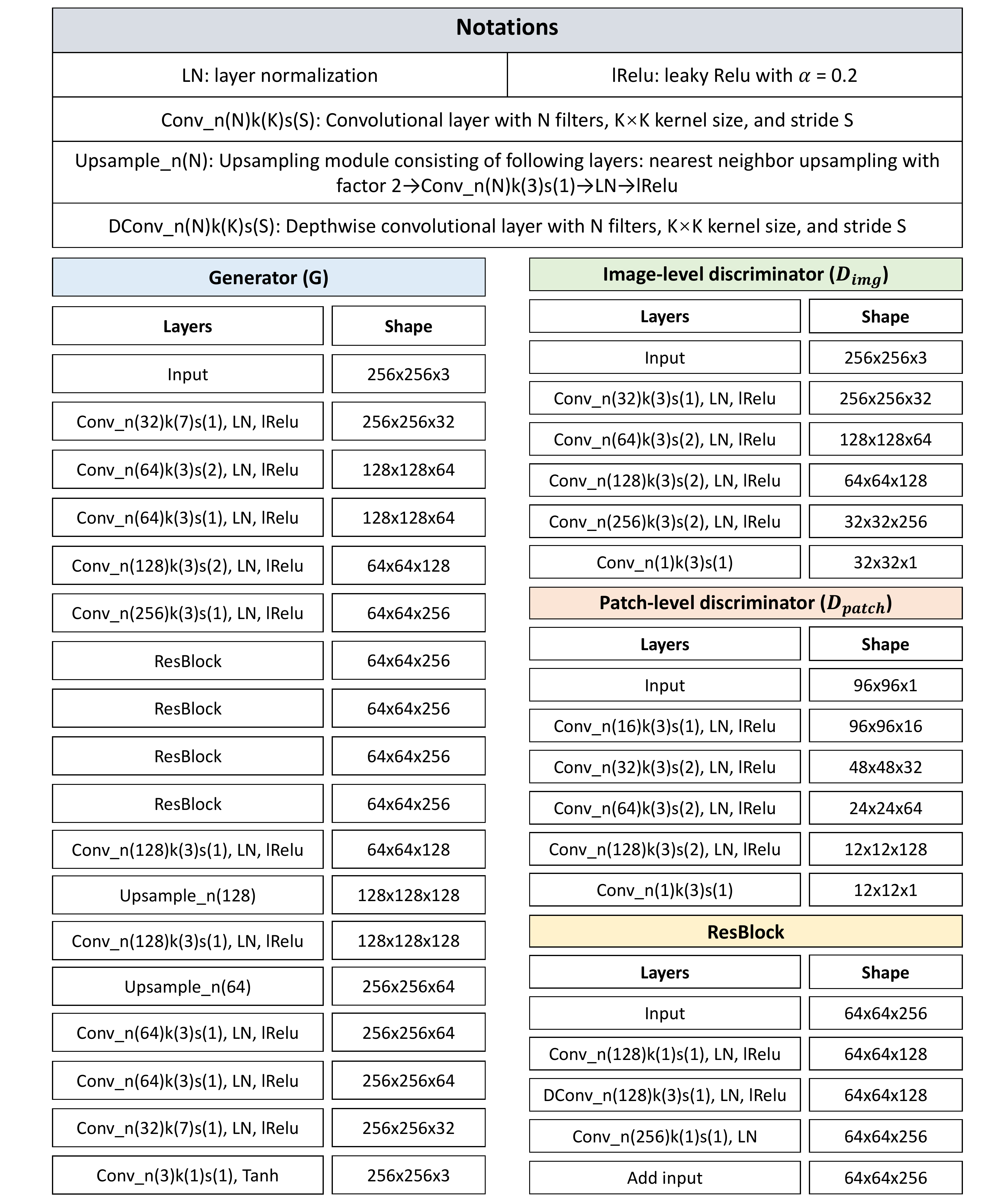}
\caption{Network details of the generator $G$, the image-level discriminator $D_{img}$, and the patch-level discriminator $D_{patch}$.}
\label{network_details}
\end{figure*}

\begin{figure*}[t]
\center
\includegraphics[width=6.5in]{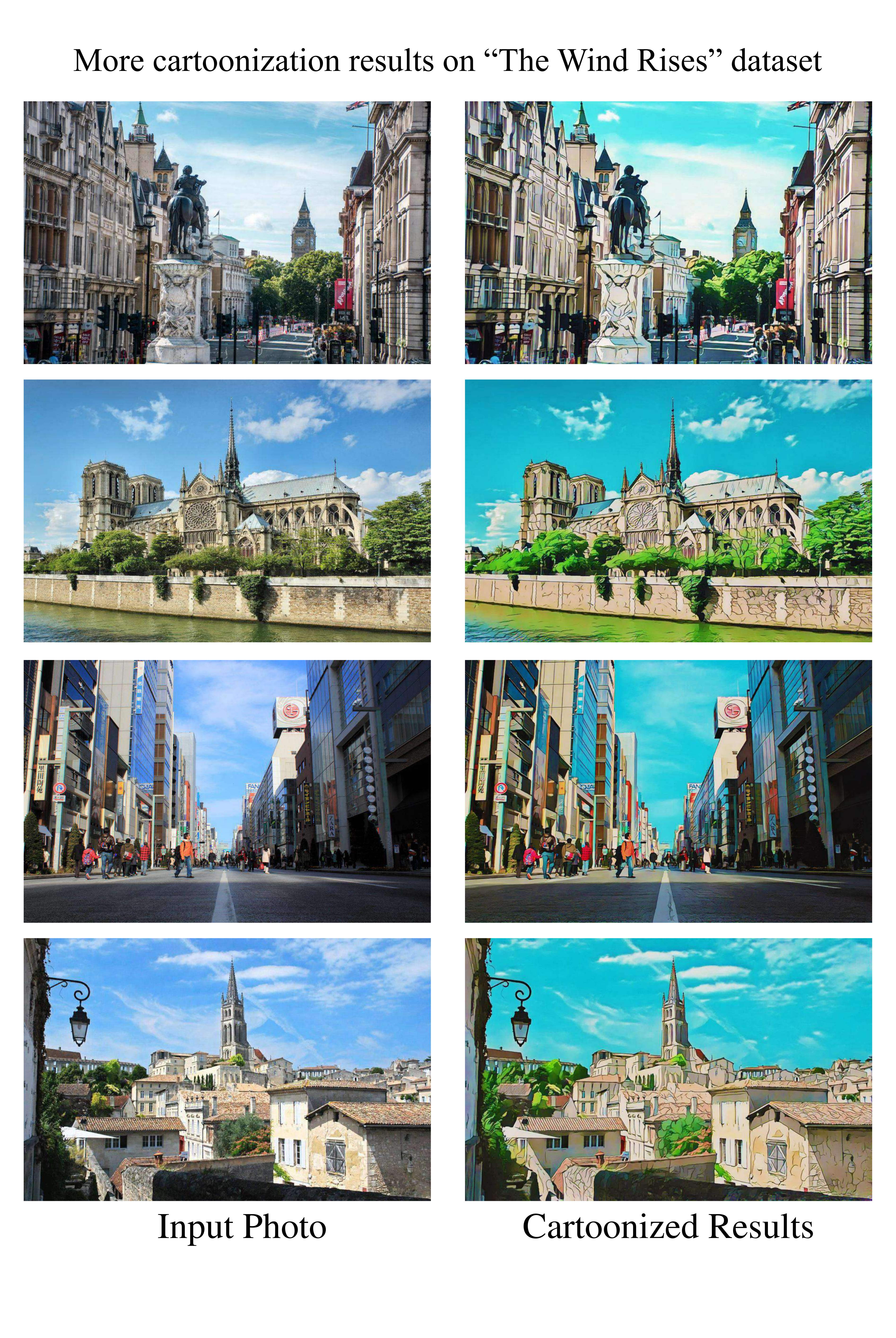}
\label{more_results_1}
\end{figure*}

\begin{figure*}[t]
\center
\includegraphics[width=6.5in]{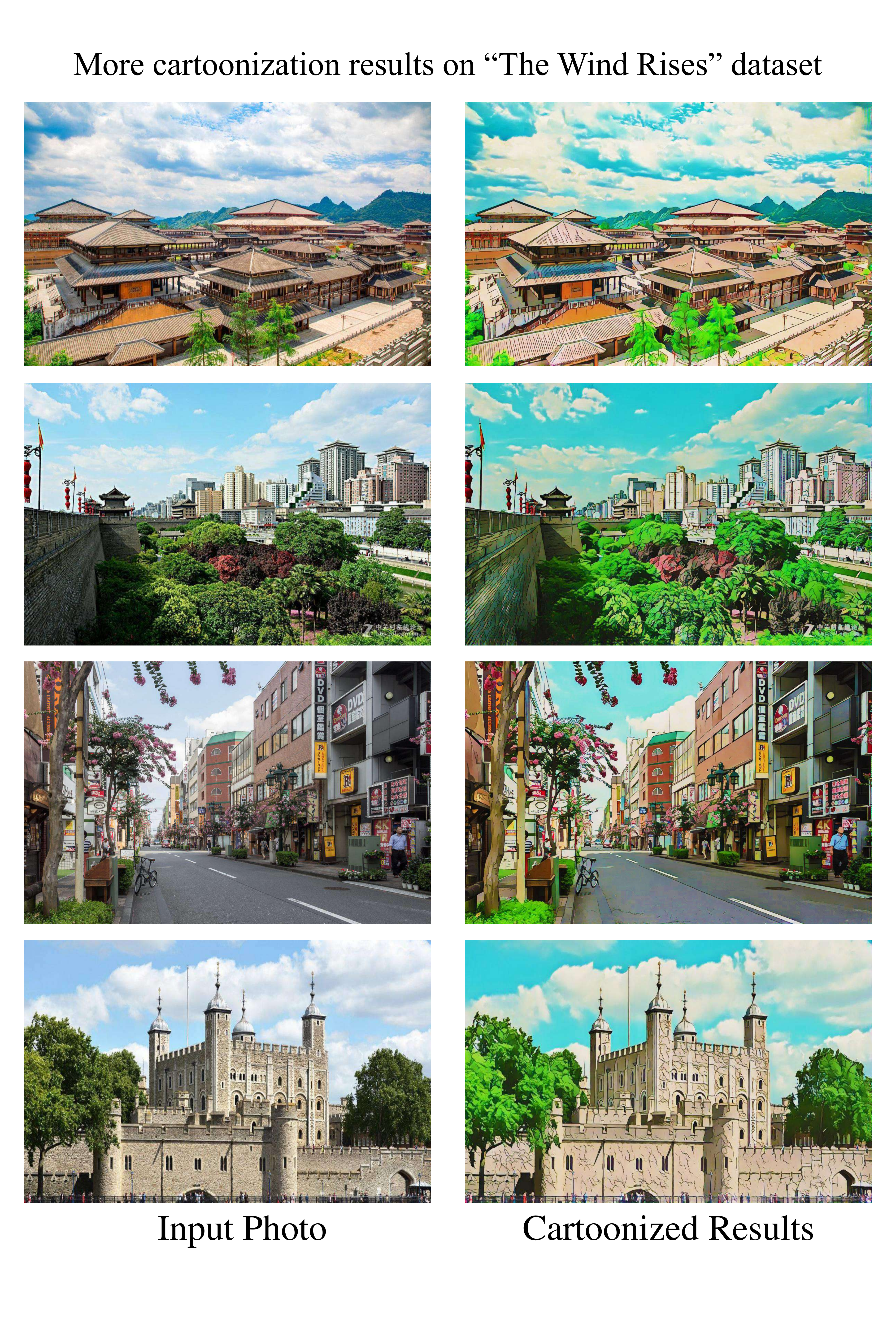}
\label{more_results_2}
\end{figure*}

\begin{figure*}[t]
\center
\includegraphics[width=6.5in]{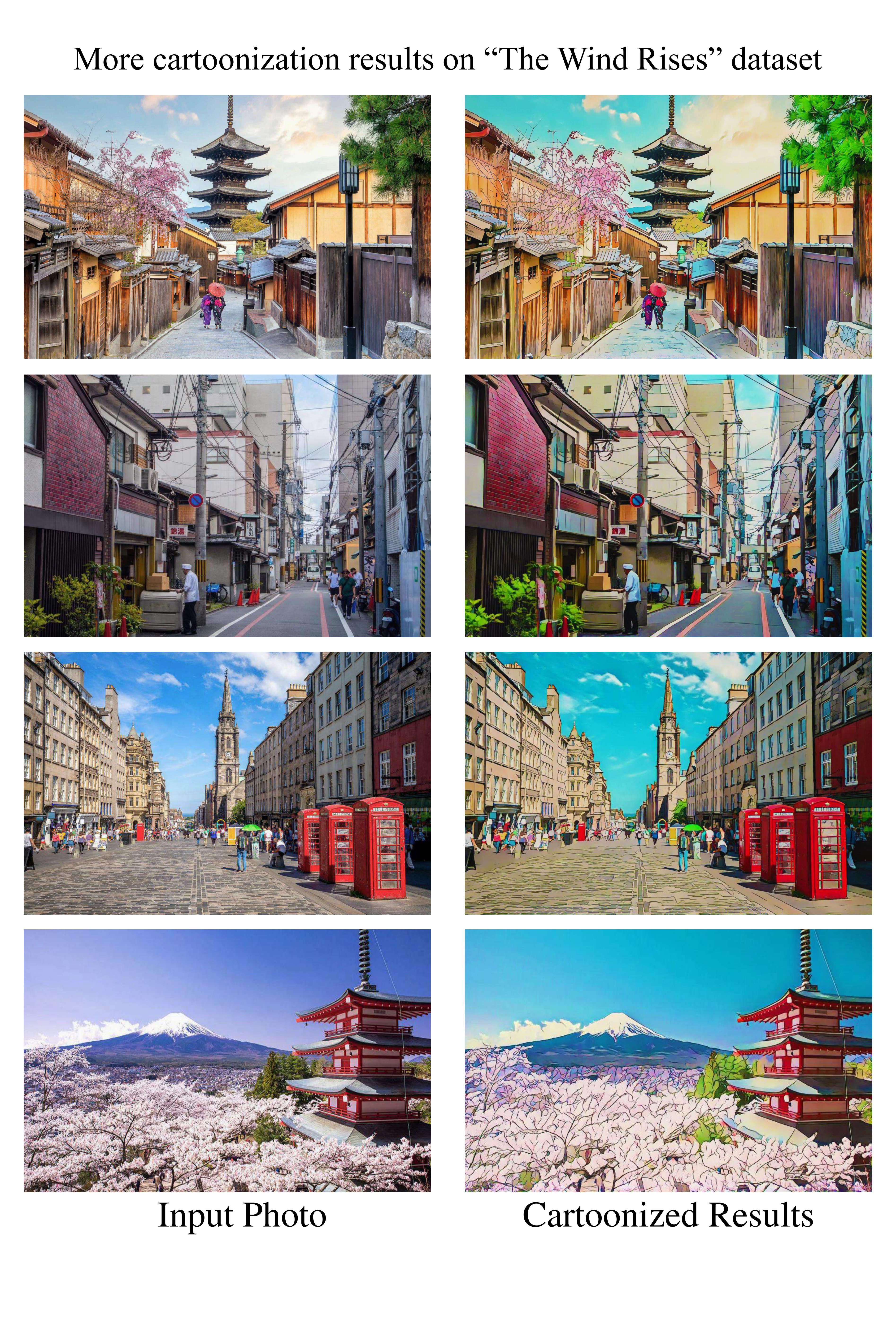}
\label{more_results_3}
\end{figure*}

\begin{figure*}[t]
\center
\includegraphics[width=6.5in]{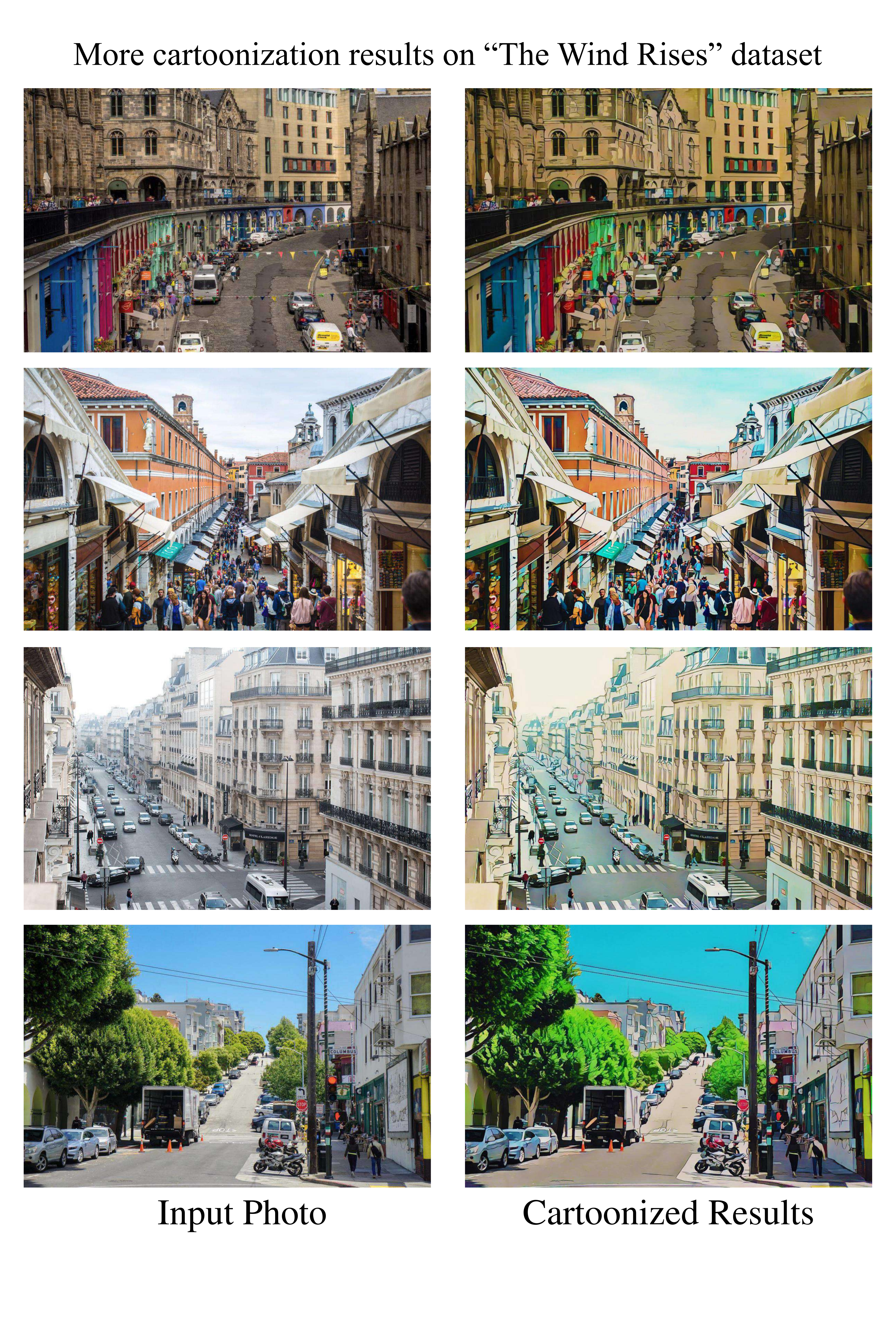}
\label{more_results_4}
\end{figure*}

\begin{figure*}[t]
\center
\includegraphics[width=6.5in]{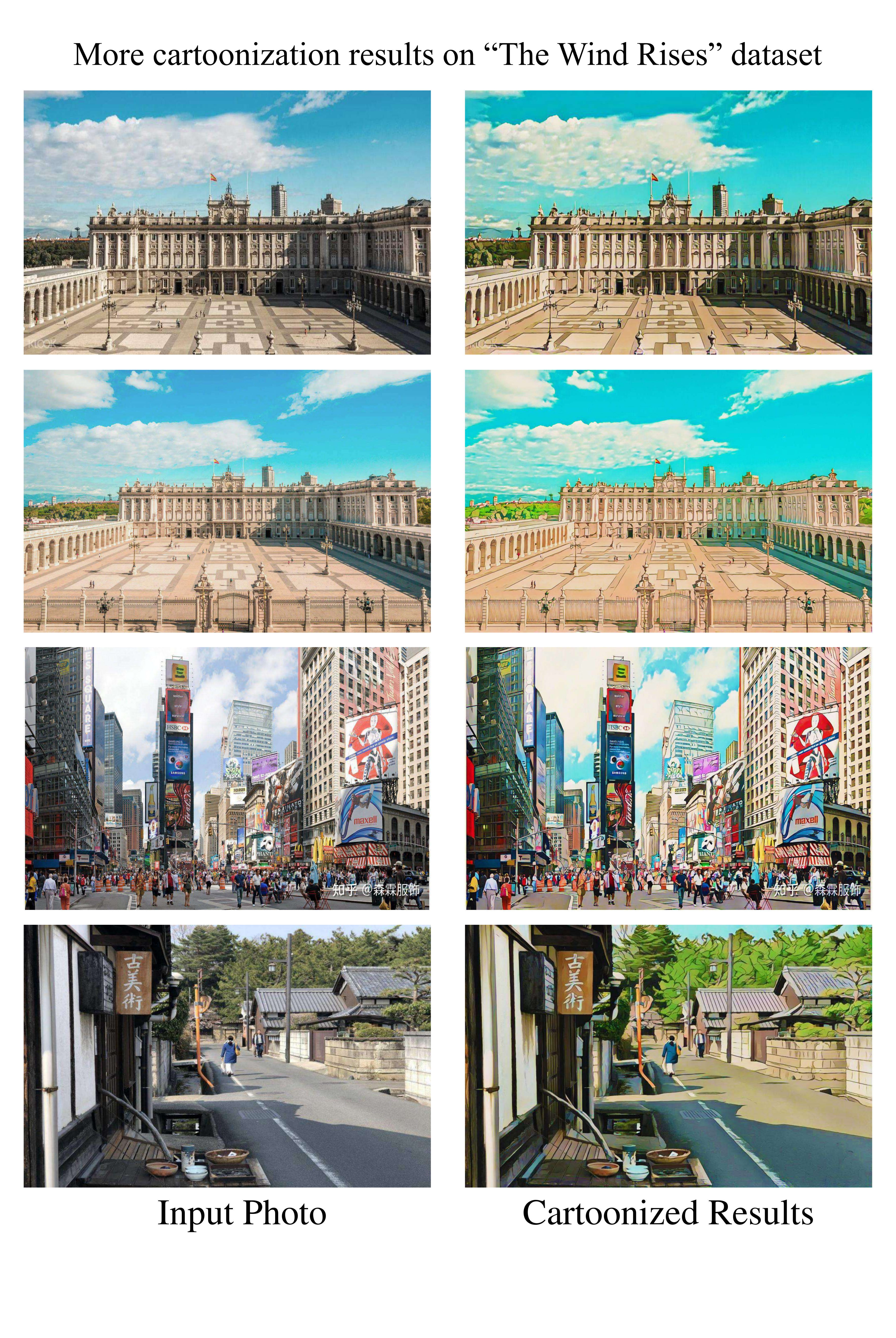}
\label{more_results_5}
\end{figure*}

\begin{figure*}[t]
\center
\includegraphics[width=6.5in]{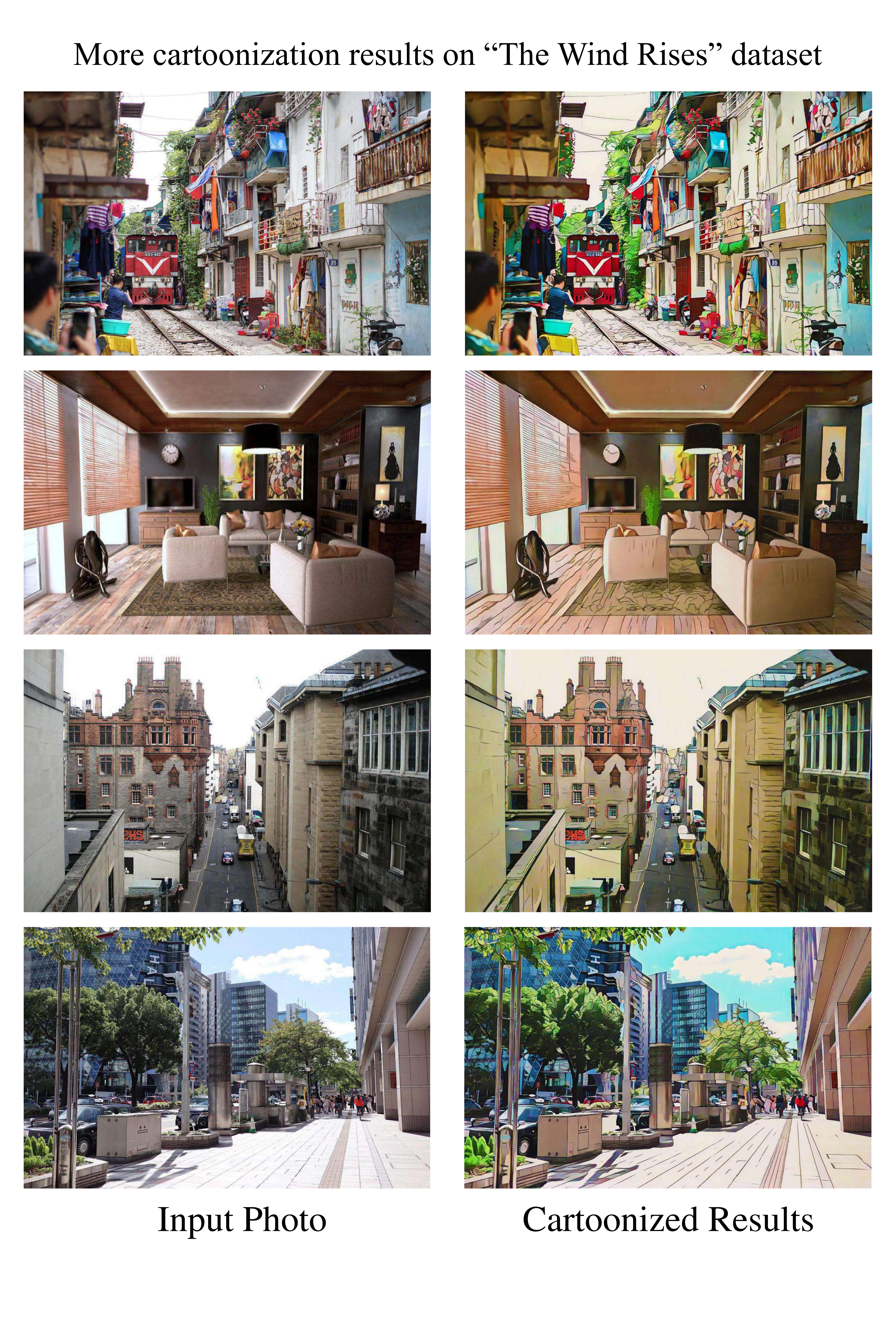}
\label{more_results_6}
\end{figure*}

\begin{figure*}[t]
\center
\includegraphics[width=6.5in]{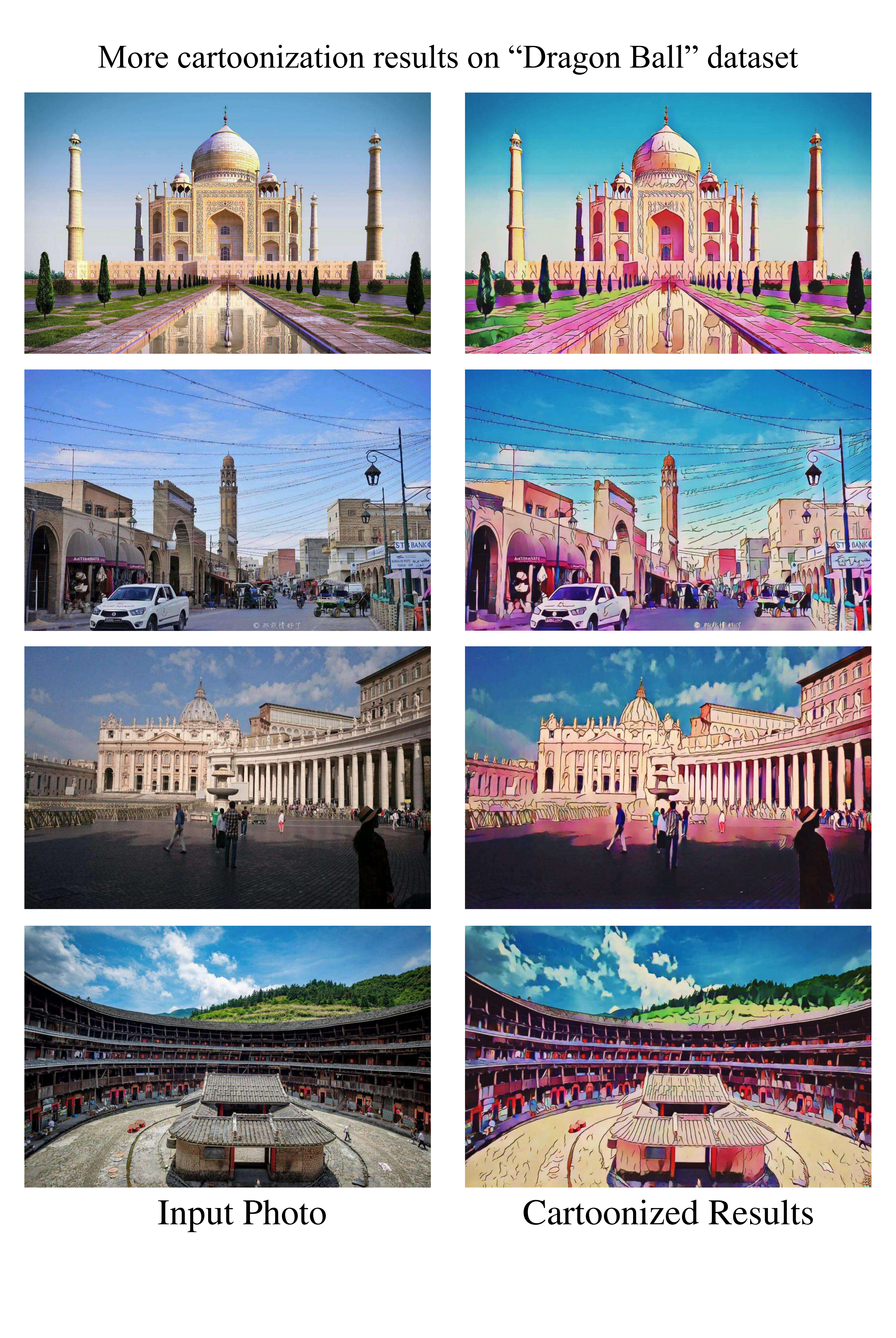}
\label{more_results_7}
\end{figure*}

\begin{figure*}[t]
\center
\includegraphics[width=6.5in]{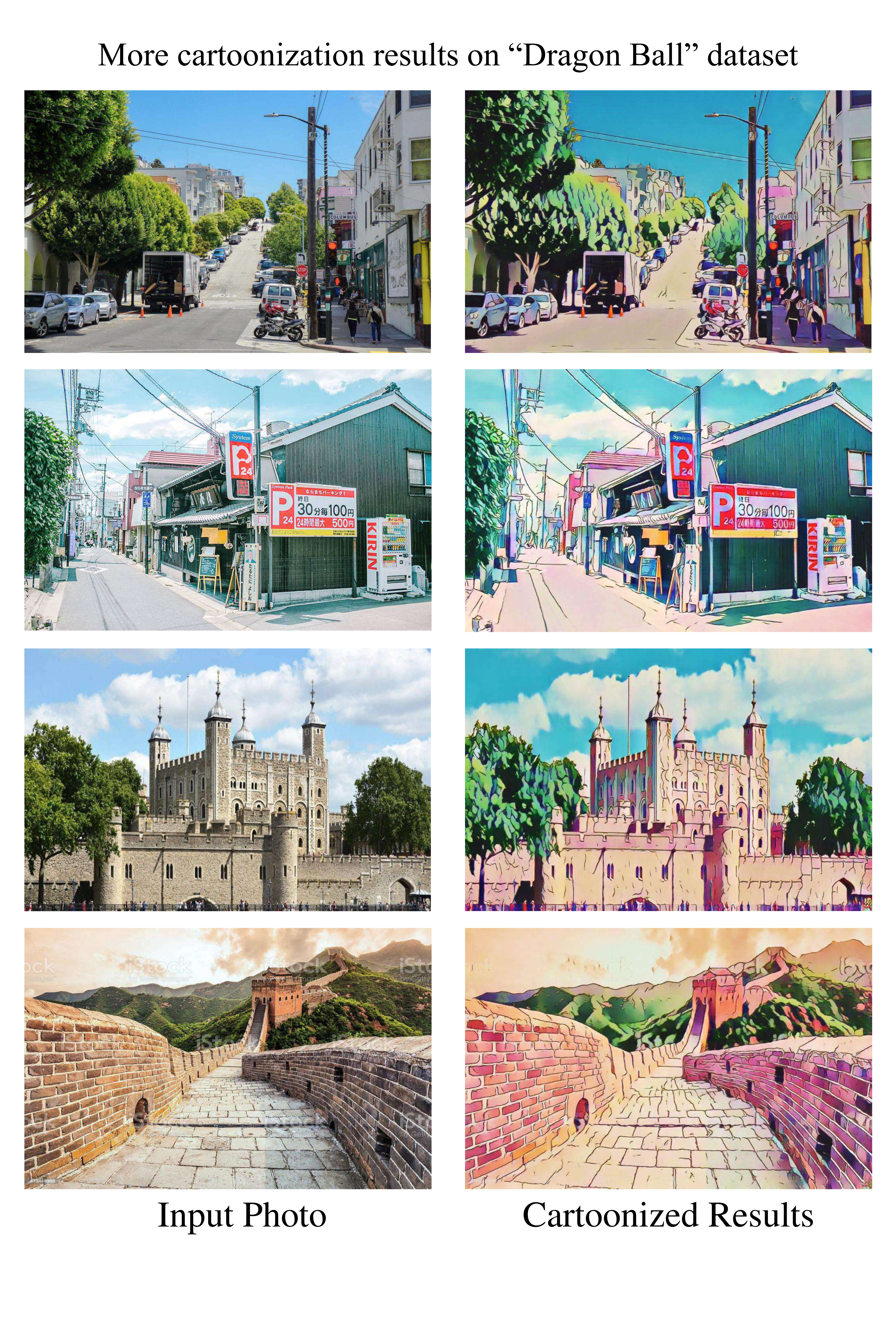}
\label{more_results_8}
\end{figure*}

\begin{figure*}[t]
\center
\includegraphics[width=6.5in]{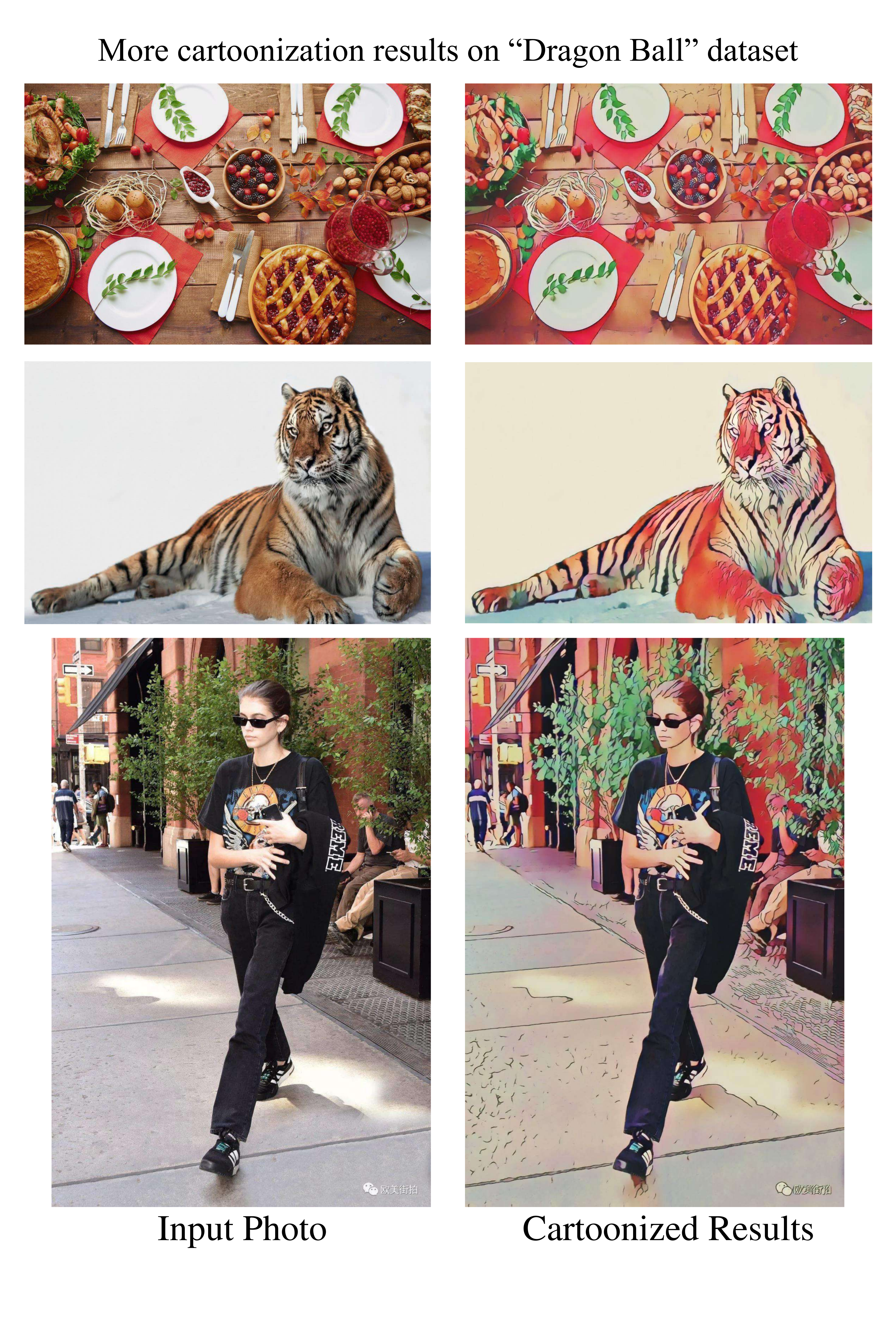}
\label{more_results_9}
\end{figure*}

\begin{figure*}[t]
\center
\includegraphics[width=6.5in]{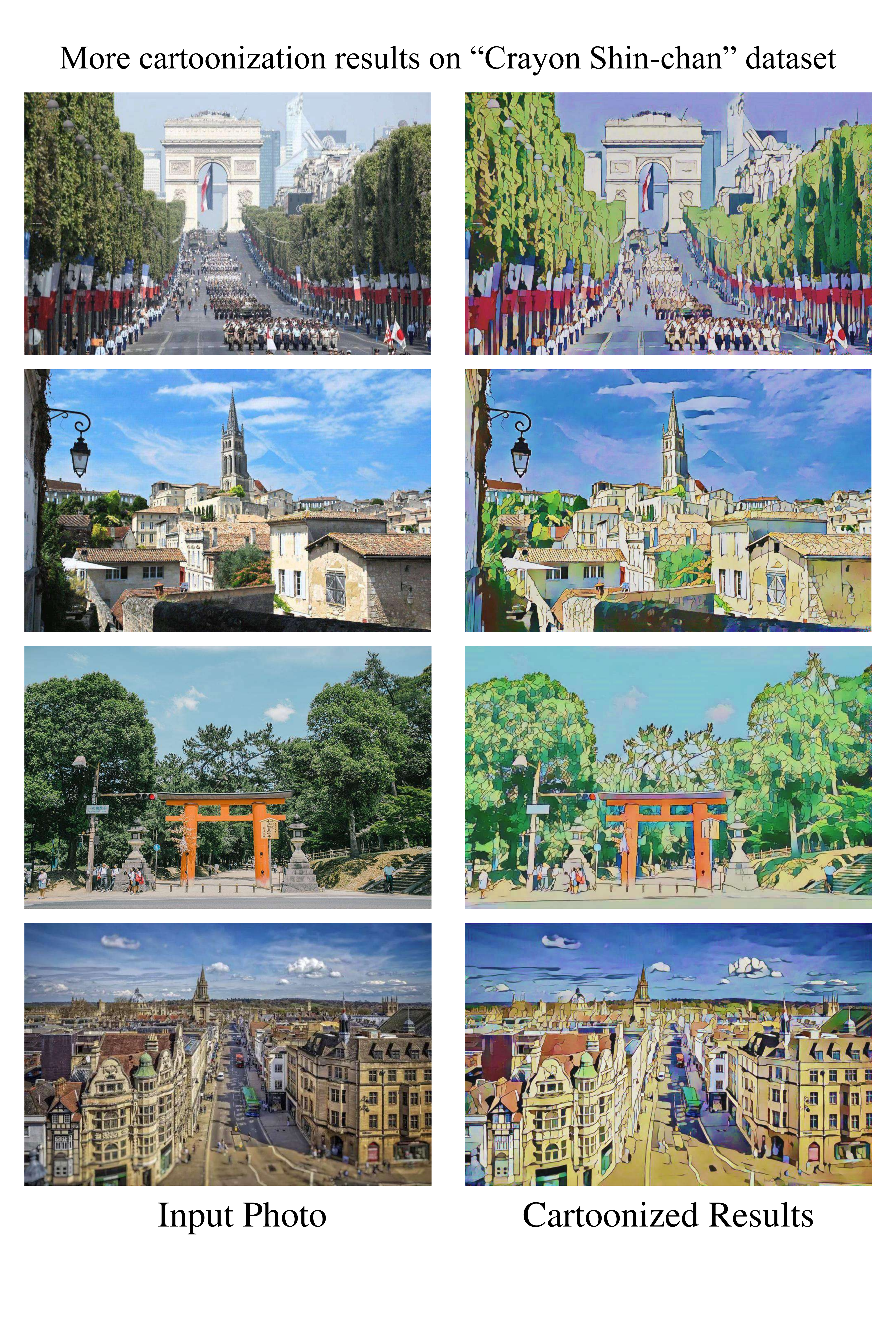}
\label{more_results_10}
\end{figure*}

\begin{figure*}[t]
\center
\includegraphics[width=6.5in]{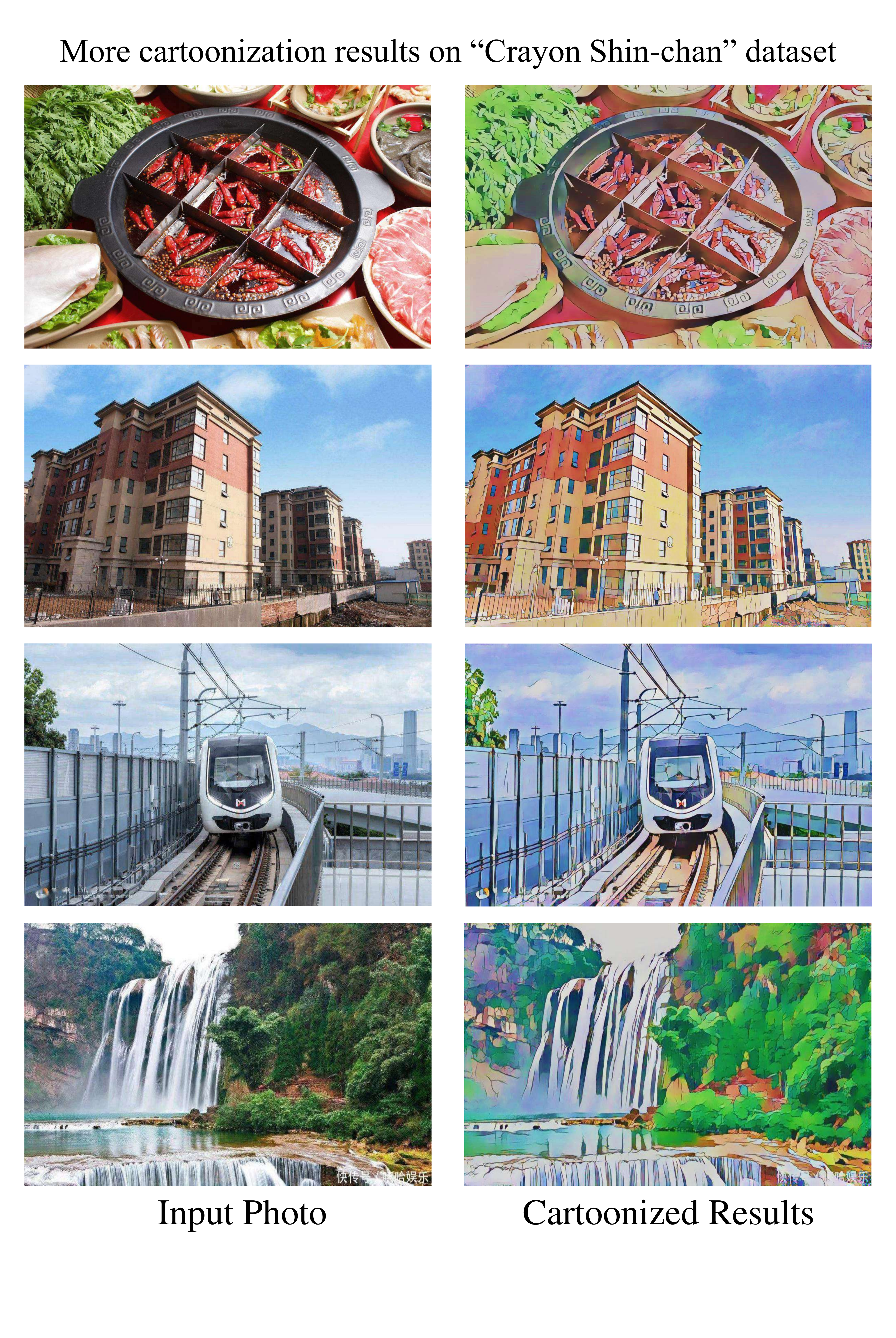}
\label{more_results_11}
\end{figure*}

\begin{figure*}[t]
\center
\includegraphics[width=6.5in]{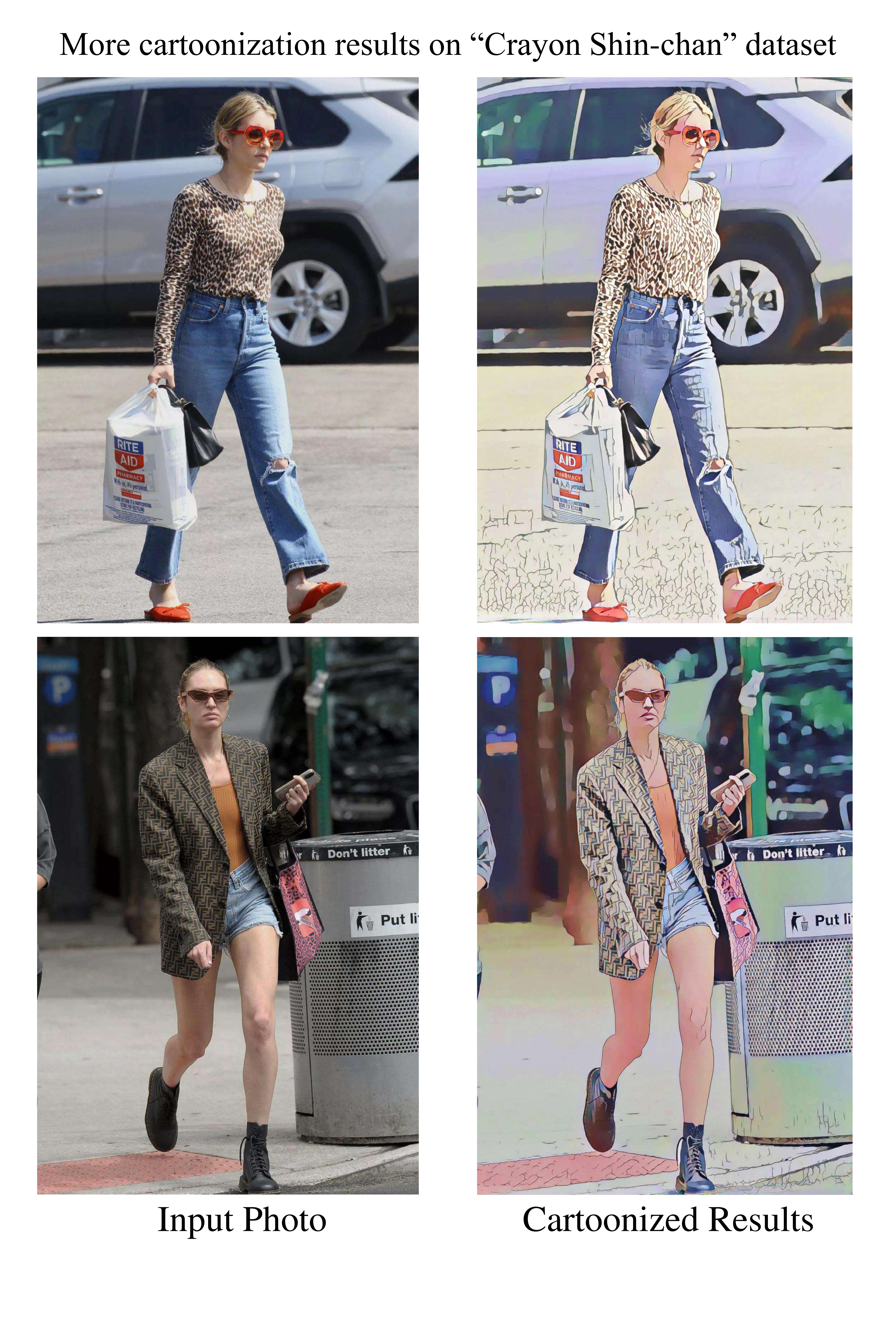}
\label{more_results_12}
\end{figure*}

\end{document}